\documentclass{article}

\usepackage{arxiv}

\usepackage[utf8]{inputenc} % allow utf-8 input
\usepackage[T1]{fontenc}    % use 8-bit T1 fonts
\usepackage{hyperref}       % hyperlinks
\usepackage{url}            % simple URL typesetting
\usepackage{booktabs}       % professional-quality tables
\usepackage{amsfonts}       % blackboard math symbols
\usepackage{nicefrac}       % compact symbols for 1/2, etc.
\usepackage{microtype}      % microtypography
\usepackage{lipsum}		% Can be removed after putting your text content
\usepackage{subcaption}
\usepackage{amsthm}
\usepackage{graphicx}
\usepackage{doi}
\usepackage{physics}
\usepackage{amssymb}% http://ctan.org/pkg/amssymb
\usepackage{pifont}% http://ctan.org/pkg/pifont

\newcommand{\pd}[2]{\frac{\partial #1}{\partial #2}}
\newcommand{\pdd}[2]{\frac{\partial^2 #1}{\partial #2^2}}
\newcommand{\bs}{\boldsymbol}
\newcommand{\mbf}{\mathbf}

\newcommand{\vdotop}{\mathbin{\vcenter{\offinterlineskip
\hbox{\(\cdot\)}
\kern -0.2ex
\hbox{\(\cdot\)}
\kern -0.2ex
\hbox{\(\cdot\)}
}}}

\usepackage{setspace}
\title{From inverse problems to neural operators: prediction, mechanism, and generalization of data-driven models}

\date{} 					% Or removing it

\author{
    Conor Rowan \\
	Aerospace Engineering\\
	University of Colorado Boulder\\
    3775 Discovery Drive \\
	Boulder, CO 80309 \\
	\texttt{conor.rowan@colorado.edu} \\
}

\renewcommand{\headeright}{}
\renewcommand{\undertitle}{}
\renewcommand{\shorttitle}{}

\begin{document}
\maketitle

\begin{abstract}

Scientists and engineers have historically relied on mathematical models based on ordinary and partial differential equations to relate system inputs---forces, fluxes, or heat sources---to outputs, such as displacement, velocity, concentration, and temperature. These models rely on deep domain knowledge to determine the form of the governing differential equation, which is then calibrated with data by solving a so-called inverse problem for material parameters, such as stiffness or heat conductivity. In recent years, the field of Scientific Machine Learning has introduced a variety of alternative modeling strategies for physical systems. A method called Sparse Identification of Nonlinear Dynamics learns the governing equation as a sparse linear combination of terms in a user-defined library. Neural Ordinary Differential Equations construct the governing equation by taking in the state and its derivatives at the input layer of a neural network. Entirely foregoing the modeling framework of differential equations, neural operators (Deep Operator Network, Fourier Neural Operator, etc.) directly learn a non-linear mapping between the system inputs and outputs. From inverse problems to neural operators, all of these modeling strategies can be conceptualized as data-driven machinery to predict a system's response over a range of inputs. Seen from this perspective, it is natural to wonder how exactly these various strategies relate to each other, and whether they can be neatly taxonomized. Drawing from the philosophical literature on scientific models, we argue that many model types have a common structure, differing only in the assumed model class of the input-output relation they define. Connecting to philosophical ideas on mechanism, and arguing that data from physical systems arises from solutions to parsimonious differential equations, we propose that only certain models are capable of mechanism discovery, and thus generalization. Our analysis is intended to unite apparently disparate modeling strategies and provide insight into their appropriate use cases.
\end{abstract}

% keywords can be removed

%------------------------------------------------------------------------------

\section{Introduction}

\paragraph{} From both technical and philosophical perspectives, this paper investigates the process of building models of physical phenomena with data. Our background is in engineering mechanics, which is concerned primarily with the application of conservation of mass, momentum, and energy to continuum systems. This encompasses the fields of solid mechanics, fluid mechanics, heat transfer, and mass transport, which are all extremely mature and well-studied fields.\footnote{Solid mechanics is the study of forces and deformation in material bodies, and is used to certify that load carrying structures like bridges or buildings are safe. Fluid mechanics is used in the context of aerodynamics to understand the flow of air around bodies, and, as an example, to design aircraft wings that maximize lift and minimize drag. Heat transfer analyzes the dynamics of heat and temperature in materials, and can be used to ensure that heated components do not melt. The field of mass transport studies the motion of mass in a background medium, and can be used to predict how far contaminants spread in media such as soil.} Accordingly, our analysis of models is informed by experience with these sorts of problems, though we anticipate that our conclusions extend beyond engineering mechanics. Our goals are to show that 1) models in engineering have a common form, 2) to argue that only certain kinds of models generalize, and 3) apply these conclusions to four modeling strategies commonly encountered in the data-driven modeling literature. Arguing that all scientific models have empirical components, and are thus data-driven in some capacity---even if only through fundamental physical constants like Planck's constant or the gravitational constant \cite{naser_what_2026}---we take a model to generalize when it makes accurate predictions beyond the scenario(s) on which it was calibrated. The notion of generalization will be clarified and discussed further in subsequent sections. In presenting our arguments, we weave together technical details of data-driven modeling with discussions in the philosophical literature on models and mechanisms.

\section{Building a model}

\paragraph{} To begin, we pose a simple question: \textit{what must a model do?} While debated in the philosophical literature, we claim that in many scientific settings, and especially in engineering, one of the primary purpose of a model is to make accurate predictions. To be sure, models can be a search for underlying truth \cite{maxwell_ontological_1962}, mechanistic explanations \cite{machamer_thinking_2000}, or provide testing grounds for scientific hypotheses \cite{rohwer_how_2016}, but we choose to view models as instruments of prediction. We claim that an emphasis on prediction over mechanism and explanation is appropriate in many physical sciences, where the modeler's work is proximate to fundamental laws of nature, which taken as brute facts, are not amenable to explanation or mechanistic analysis \cite{kitcher_explanatory_1981}. In other words, laws of nature (e.g., conservation laws) can be used to explain, but are not themselves explainable \cite{mckenna_laws_2026}. We also restrict attention to mathematical and computational models, rather than physical scale models, such as the San Francisco Bay model discussed in \cite{weisberg_simulation_2013}. In the context of physical sciences, we can be very specific about the form of predictive machinery furnished by a mathematical model. Broadly speaking, we take a model $\mathcal M$ to be a map between an input field $f(\mbf x)$ and output field $u(\mbf x)$, where $\mbf x\in \Omega$ is the spatial coordinate and $\Omega$ is the domain of interest. Here, we focus on ``static'' models, meaning that there is no time dependence. Physically, static problems are infinite time limits of dynamic problems, meaning that the system has equilibrated to a steady-state---this restriction to static models facilitates analysis, but does not meaningfully limit our conclusions. To align with more colloquial usage in the data-driven modeling literature, we refer to $u(\mbf x)$ as the system ``state,'' and $f(\mbf x)$ as the ``forcing.'' In engineering mechanics, the state is often a displacement, velocity, temperature, or concentration field, whereas the forcing is a force, heat flux, or mass input. With all this in mind, we write the model abstractly as

\begin{equation*}
    \mathcal{M} : f(\mbf x) \mapsto u(\mbf x), \quad \mbf x\in\Omega.
\end{equation*}

\begin{figure}[hbt!]
\centering
\includegraphics[width=0.99\textwidth]{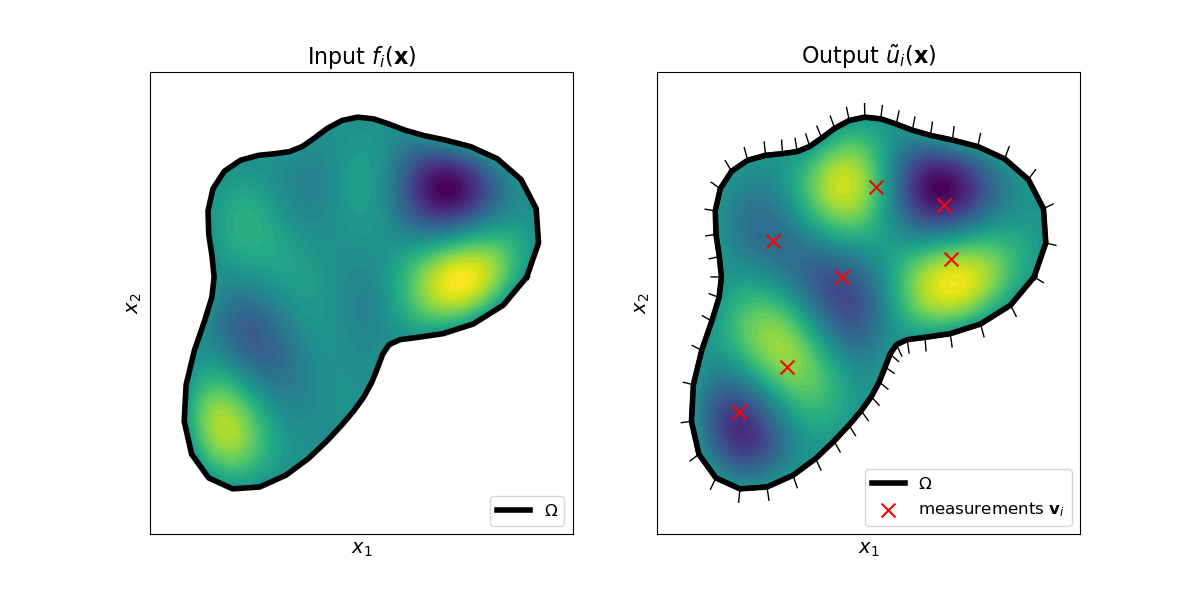}
\caption{A schematic of a system of interest and its experimental characterization. We take the goal of model building to be a map relating the input and output fields. Equipped with such a map, the modeler is able to make novel predictions about the system's response. We distinguish between the true system response $\tilde u(\mbf x)$ and the model of that response $u(\mbf x)$. Our domain of interest is a ``space potato,'' inspired by the geometry of example problems in texts on continuum mechanics \cite{reddy_introduction_2013}.}
\label{potato}
\end{figure}

The goal of model-building is to learn the map $\mathcal M$ between input forcings and output states based on a finite ``training'' data set. The training data is denoted as $\mathcal D = \{ f_i(\mbf x) , \mbf v_i \}_{i=1}^N$, where each experiment, indexed by $i$, corresponds to a particular continuous input forcing $f_i(\mbf x)$, and an observation of the system state at a discrete set of measurement points, which we call $\mbf v_i$. The total number of experiments is $N$. This is shown schematically in Figure \ref{potato}. Put simply, the experimentalist probes the system with input fields of their choosing, and observes the response at a discrete set of points, perhaps corresponding to sensor positions. We take the input to be known continuously in space, as opposed to at a discrete set of sensor positions, as this is something the experimentalist can explicitly control, though we do not think this simplifying assumption is restrictive. We denote the number of measurements as $\mathcal S$, such that each measurement is a vector $\mbf v_i \in \mathbb R^{\mathcal S}$. As it currently stands, the model is a map between functions, a mathematical object called an operator. In order to formulate a model building problem that can be implemented in digital computers, we must somehow ``discretize'' the operator, meaning that we seek a finite dimensional representation of $\mathcal M$. In particular, our model will be constructed with a finite set of parameters $\bs \theta$. To understand how these parameters appear, we write both the input and output fields with a set of basis functions given by $\{h_j(\mbf x)\}_{j=1}^{\mathcal B}$, where $\mathcal B$ is the dimension of the discretization. Thus, the input and output fields are discretized with the basis as 

\begin{equation*}
    f(\mbf x) = \sum_{j=1}^{\mathcal B} F_j h_j(\mbf x), \quad u(\mbf x) = \sum_{j=1}^{\mathcal B} U_j h_j(\mbf x).
\end{equation*}

\begin{figure}[hbt!]
\centering
\includegraphics[width=0.99\textwidth]{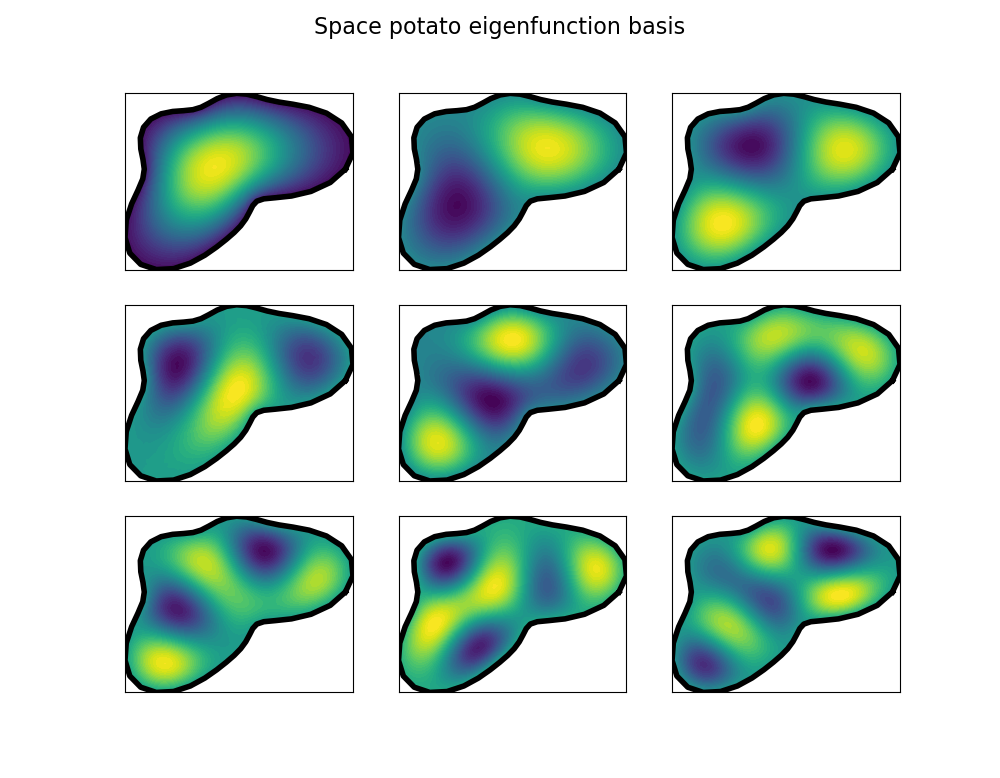}
\caption{A set of $9$ Laplace eigenfunctions on the space potato domain which can be used to discretize the forcing and state fields. Discretizing the input and output spaces also allow for a finite dimensional representation of the model.}
\label{eigenfunctions}
\end{figure}

\noindent where $\mbf F \in \mathbb R^{\mathcal B}$ provides the discrete representation of the input forcing, and $\mbf U \in \mathbb{R}^{\mathcal B}$ the discrete representation of the state. Typically, basis functions are chosen such that a large class of functions can be accurately approximated without $\mathcal B$ being prohibitively large \cite{vallet_spectral_2008, hughes_thomas_j_r_finite_2000}. An example is so-called ``Laplace eigenfunctions,'' which are a generalization of Fourier series to arbitrary dimensions and non-square domains \cite{rowan_solving_2025}. Nine example basis functions for the domain of interest are shown in Figure \ref{eigenfunctions}. With the discretization of the input and output fields, the model then becomes a finite-dimensional function $\mbf M: \mathbb R^{\mathcal B} \rightarrow \mathbb R^{\mathcal B}$, which maps between input and output coefficients. As such, we refer to the discretized model $\mbf M$ as the ``coefficient map.'' Thus, we can write the parameterized model as

\begin{equation}\label{model}
    u(\mbf x) = \sum_{j=1}^{\mathcal B} M_j( \mbf F; \bs \theta) h_j(\mbf x),
\end{equation}

Though this choice is not unique, we will work exclusively with the discretization of the operator $\mathcal M$, and thus the model, relying on a fixed set of basis functions given by Eq. \eqref{model}. We note that this choice aligns with popular numerical methods such as finite element and spectral methods, as well many other modeling strategies \cite{vallet_spectral_2008, hughes_thomas_j_r_finite_2000, li_fourier_2021, ingebrand_basis--basis_2024, tripura_wavelet_2023, lu_comprehensive_2022, willcox_balanced_2002, ghanem_stochastic_1991}. We note that modeling strategies which use neural network representations of the state do not conform to this framework \cite{raissi_physics-informed_2019, seidman_nomad_2022}, and finite difference methods also do not employ a fixed basis. Though not entirely general, we take the fixed basis representation of the model to be a helpful tool to unify disparate modeling strategies. One reason we have chosen to focus on static problems is because this discretization of the problem in terms of a basis is more common and natural. A summary of notation introduced thus far is given in Table \ref{tab:notation}.

\renewcommand{\arraystretch}{1.5}
\begin{table}[h]
    \centering
    \caption{A summary of notation used in setting up the problem of building a data-driven model using a fixed basis to represent the input and output fields.}
    \label{tab:notation}
    \begin{tabular}{|c|c|}
        \hline
        \textbf{Symbol} & \textbf{Name}  \\
        \hline
        $u(\mbf x)$ & Output/state  \\ \hline
        $f(\mbf x)$ & Input/forcing \\ \hline
        $\mathcal M$ & Abstract model  \\ \hline
        $\Omega$ & Domain geometry \\ \hline
        $\mbf v_i$ & Measurement data ($i=1,\dots,N$ ) \\ \hline
        $h_j(\mbf x)$ & Basis functions ($j=1,\dots,\mathcal B$)   \\ \hline
        $\mbf U$ & Discretized state  \\ \hline
        $\mbf F$ & Discretized forcing  \\ \hline
        $\mbf M$ & Discretized model/coefficient map  \\ 
        \hline
        $\bs \theta$ & Model parameters \\ \hline
    \end{tabular}
\end{table}

\paragraph{} We note that we have given a very minimal account of scientific modeling. In particular, we have concerned ourselves only with models as prediction machines with an overarching mathematical structure of Eq. \eqref{model}. This choice sidesteps many of the philosophical questions about scientific models which are discussed in the philosophy literature, which is necessary at this point, given that we are intentionally vague about the details of the model. But before proceeding any further, we situate our approach in the context of these philosophical discussions. First, we follow a number of authors in seeing a primary use of models as the application of fundamental laws of nature to particular systems of interest \cite{winsberg_science_2010, herfel_theories_2023, morgan_models_1999}. Like Morrison, we see models as ``mediators between theory and application,'' as laws do not come ready-made with instructions for how they should be applied to specific systems \cite{morrison_reconstructing_2015}. For example, continuum mechanics is a broad and active field of research, but is ultimately the study of the application of well-known conservation laws. However, not all models make use of fundamental physical laws. At this point, taking our modeling efforts to be encapsulated entirely in the coefficient map $\mbf M(\mbf F; \bs \theta)$, we do not distinguish between phenomenological and causal/mechanistic/explanatory models. McMullin defines a phenomenological model as an ``arbitrarily chosen mathematically-expressed correlation of physical parameters from which the empirical laws of some domain can be derived'' \cite{mcmullin_what_1968}. While phenomenological models are mere summaries of data \cite{termine_machine_2026}, models with causal, mechanistic, and/or explanatory content are considered to have some property beyond mere summarization. For many authors, this property is ``representation,'' meaning that the model captures real features of the target system in some way \cite{giere_how_2004, frigg_models_2021, morgan_models_1999}. However, rigorously defining representation in the context of science has been challenging, with entire works specifically devoted to this task \cite{van_fraassen_imaging_2008, nguyen_scientific_2022}. As an alternative to representation, the idea of ``mechanism'' has been proposed as a necessary ingredient of scientific models which do more than summarize. Machamer, Darden, and Craver define a mechanism as ``entities and activities organized such that they are productive of regular changes from start or set-up to finish or termination conditions'' \cite{machamer_thinking_2000}. Definitions of mechanism which rely on decomposing a system into parts (e.g., entities and activities), have been popular in the so-called ``New Mechanist'' literature \cite{craver_when_2006, siegel_phenomenological_2024, povich_mechanistic_2017}. In the context of engineering, fundamental physical laws and mechanisms often go hand-in-hand. For example, models for the dynamics of temperature in material bodies are based on conservation of energy (fundamental law), but rely on specific assumptions about how energy is transported in the body (mechanisms). 

\paragraph{} Ultimately, we are concerned more with the methodology of model building than its ontology and/or metaphysics \cite{ioannidis_mechanisms_2018}. As such, the philosophical stance we adopt is akin to Suarez's ``inferential'' conception of models, whereby a structure (computational, mathematical, or physical) becomes a model by allowing the modeler to draw inferences about the target system \cite{suarez_inferential_2004}. Subsequent inferentialists extend Suarez's account specifically to applied mathematics, arguing that ``the fundamental role of mathematics is inferential: by embedding certain features of the empirical world into a mathematical structure, it is possible to obtain inferences that would otherwise be extraordinary hard to obtain'' \cite{bueno_inferential_2011}. Another modification of the inferentialist account of scientific models is given in \cite{khalifa_scientific_2022}, where the authors address the problem of scientific representation by claiming a model represents when it allows the modeler to draw inferences about the target system. These authors thus emphasize the ability of the model to \textit{predict} the behavior of the target system, thus allowing the modeler to draw inferences. For example, the ability of a model to make predictions allows the modelers to infer system behavior under a wide range of operating conditions. Along similar lines, the interventionist approach to models claims that the model should allow the modeler to reason about the behavior of the system when inputs are changed, but interventionists emphasize more strongly the role of causal relationships in the modeling process \cite{woodward_what_2002}. As Woodward notes, the model may make meaningful predictions ``only over a limited spatio-temporal interval'' or may have ``exceptions or narrow scope.'' For the time being, we adopt neither a strictly inferentialist nor an interventionist stance on models, deferring a discussion of the role that mechanism and causality play in making predictions to later. At this point, we interest ourselves in the relationship between the mathematical form of the parameterized model $\mbf M(\mbf F; \bs \theta)$ and its scope, meaning its ability to generalize beyond the training data set. In other words, having argued that models have a common form, and thus seeing that the model is determined entirely by the coefficient map, we now ask: what is the difference between coefficient maps with narrow scope---those that merely summarize data---and those with broad scope, and thus the ability to generalize? By generalization, we mean that the model makes accurate predictions beyond the scenarios used to calibrate it. We will see that answering this question ultimately brings us back to philosophical questions of mechanism and causality. However, before answering this question, we first must lay out how the training data is generated and collected, and then clearly define a notion of generalization.

\section{Data generating process}

\paragraph{} Figure \ref{potato} is a schematic of the data collection process: the experimentalist probes the system with an input forcing $f_i(\mbf x)$, the system's steady-state response is $\tilde u_i(\mbf x)$, but the experimentalist only has access to this response through a discrete set of measurements $\mbf v_i \in \mathbb R^{\mathcal S}$. Now, we make an assumption that we take to be unobjectionable for many scientific fields: the system of interest obeys a partial differential equation (PDE). In other words, the true input-output map $f(\mbf x) \mapsto \tilde u(\mbf x) $ is given by the solution of some PDE. For the sake of illustration, we consider a reaction-diffusion PDE. The governing equation for the state field is

\begin{equation}\label{radiation}
\begin{aligned}
    \underbrace{\kappa \nabla^2 \tilde u(\mbf x)}_{\text{diffusion}} + \underbrace{\sigma \tilde u(\mbf x)(1-\tilde u(\mbf x)^2)}_{\text{reaction}} + \underbrace{f(\mbf x)}_{\text{mass source}}=0, \quad \mbf x \in \Omega ,\\
    \underbrace{\tilde u(\mbf x) =0}_{\text{boundary condition}} , \quad \mbf x \in \partial \Omega.
\end{aligned}
\end{equation}

To reiterate, notation $\tilde u$ is used to distinguish the true output of the system from the model prediction $u$. Eq. \eqref{radiation} is a steady-state Allen-Cahn equation, a prototypical 
reaction-diffusion model for phase separation of binary media \cite{bretin_penalized_2024}. The notation $\partial \Omega$ denotes the boundary of the system, and meaning of the various terms in the governing PDE are labeled for clarity. The diffusion term encourages the state field to be a spread out version of the source, and the reaction term pushes the state to be either $0$ or $1$. The zero boundary condition is chosen for simplicity, but, again, we do not think this meaningfully limits the generality of our discussion.\footnote{The hashed boundary in Figure \ref{potato} is used to indicate the zero boundary condition.} We believe it is both simpler and more illuminating to talk about specific problems rather than abstract ones. This model is parameterized by two material properties, the diffusivity $\kappa$, which controls the rate of mass diffusion, and the reaction rate $\sigma$, which controls the strength of the push to binary phases. As is standard in engineering, we view the material properties $\kappa$ and $\sigma$ as bona fide empirical parameters, though they may be derivable from kinetic theory \cite{irving_statistical_1950}. This ensures that even first-principle-based PDEs are data-driven in the sense that they must be calibrated to the system of interest through appropriate selection of material properties. This choice aligns with the idea, put forth in the philosophical literature, that fundamental mechanisms are field-specific \cite{woodward_what_2002, machamer_thinking_2000}. In our view, that one scientist's empirical model parameter is the output of another's calculation does not cheapen the claim that the model is data-driven.

\begin{figure}[hbt!]
\centering
\includegraphics[width=0.4\textwidth]{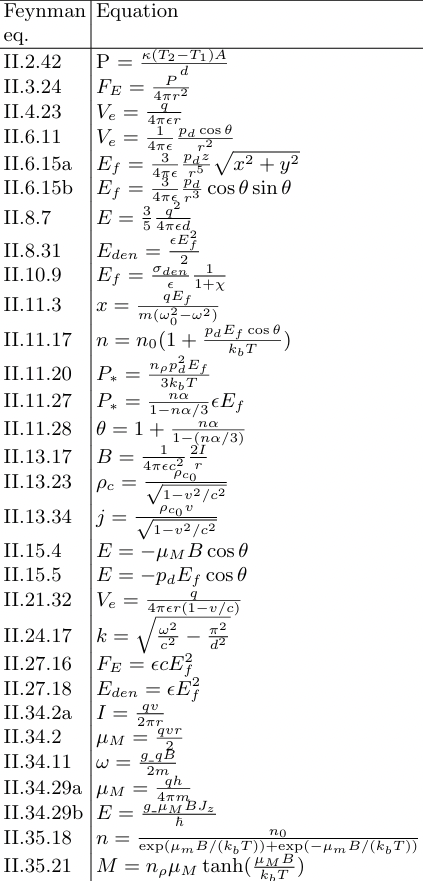}
\caption{A set of algebraic relations appearing in the Feynman lectures on physics. Adapted from \cite{udrescu_ai_2020}.}
\label{feynman}
\end{figure}

\begin{figure}[hbt!]
\centering
\includegraphics[width=0.75\textwidth]{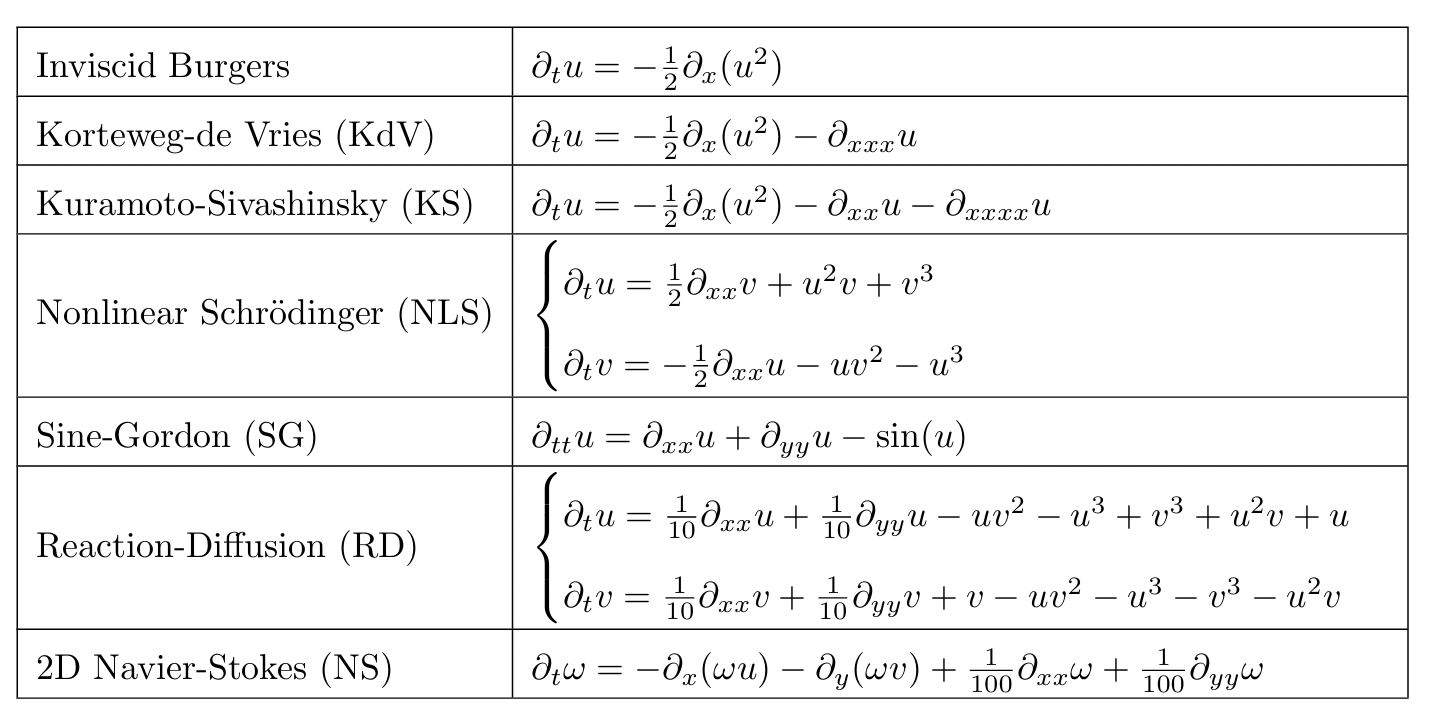}
\caption{A set of nonlinear PDEs commonly encountered in the engineering literature. Adapted from \cite{messenger_weak_2021}.}
\label{messenger}
\end{figure}

\paragraph{} In addition to assuming that the system of interest is governed by a PDE, it is also common practice in the scientific community to assume that the PDE has a particular mathematical form. For example, the equation are generally taken to be parsimonious, comprising only a small number of terms, and to have certain types of nonlinearities, meaning there are no exotic terms like $\log(\tanh( \exp(  u \nabla u)^2 ))$. This intuition has been codified in the equation discovery and symbolic regression literature, where researchers are comfortable making assumptions about the general form of equations before attempting to discover them from data \cite{brunton_discovering_2016, messenger_weak_2021, udrescu_ai_2020, cranmer_interpretable_2023}. As physicist Eugene Wigner says, that our mathematics so elegantly handles regularities of the natural world ``border[s] on the mysterious'' \cite{wigner_unreasonable_1960}. We take it as an empirical fact that the true governing equation of the system of interest is both parsimonious and mathematically simple. We note that expediency may play a role in the observed simplicity of many PDEs, as simplified models are easier to work with, yet may remain empirically adequate \cite{fraassen_arguments_1998}. While it is difficult to give a precise definition of mathematical simplicity, it suffices for our purposes to say that an expression is NOT simple when it comprises compositions of nonlinear functions. Figures \ref{feynman} and \ref{messenger} show algebraic and differential equations commonly encountered in the physical sciences, none of which include compositions of nonlinear functions. The steady-state Allen-Cahn equation is an example of a nonlinear PDE which is parsimonious and simple in our intended sense.

\paragraph{} Because most PDEs do not admit exact solutions, it is necessary to approximate solutions using computational techniques. One way of doing this is to discretize the output field with a set of basis functions, and determine the coefficients such that the approximation error is minimal in some sense. One way of accomplishing this is the Galerkin weak form, the details of which are shown in Appendix \ref{sec: allen-cahn}. In the case of the Allen-CAhn equation, the weak form gives rise to a nonlinear system of equations for the solution coefficients:

\begin{equation}\label{data_generating}
    \tilde u(\mbf x) = \sum_{j=1}^{\mathcal B} \tilde U_j h_j(\mbf x), \quad \mbf {\tilde U}(\mbf F) = \underset{\mbf{V}}{\text{solve}}\Big( -\kappa \mbf K \mbf{V} +\sigma \mbf A \mbf V - \sigma \mbf R \vdotop  ( \mbf{V} \otimes \mbf{V} \otimes \mbf{V}) + \mbf F = \mbf 0\Big).
\end{equation}

We introduce the notation ``$\text{solve}(\cdot)$'' as a concise way of writing that the output coefficients are a function of the discretized forcing $\mbf F$ through the solution of a system of equations. Furthermore, the notation ``$\vdotop$'' denotes contraction on the last three indices of the four index tensor $\mbf R$, which comes the discretization of the cubic term in the PDE. Note that while we show the particular weak form corresponding to the Allen-Cahn problem, in general, the assumption that the PDE is parsimonious and simple restricts $\mbf{\tilde U}(\mbf F)$ to be a small subset of all nonlinear functions $\mathbb R^{\mathcal B} \rightarrow \mathbb{R}^{\mathcal B}$. It is an interesting question whether nonlinear solves of discretized PDEs have something like a ``universal approximation'' property, meaning that any nonlinear function can be approximated as the discretized solution to an appropriately defined PDE. This question is very much out of scope for our current work, and we simply assume that this is not the case given our constraints of parsimony and simplicity.

\paragraph{} We do not concern ourselves with the details of how system of equations in Eq. \eqref{data_generating} is solved---it suffices to say that standard iterative techniques like Newton's method and its variants could be used to obtain a solution. Previously, we posited that a PDE governed the response of the system, and gave Eq. \eqref{radiation} as an example to make the discussion more concrete. Because we cannot solve most PDEs exactly, we must resort to numerical methods, which approximate the solution. Thus, as is common in the scientific machine learning literature, we use a numerical solution as a surrogate for a real system of interest \cite{brunton_discovering_2016, raissi_physics-informed_2019}. Our modeling problem becomes about capturing the input-output relation defined by a numerical solution to a PDE, rather than a real physical system. The extent to which these two problems are equivalent depends on the accuracy of the PDE model. In our case, we are comfortable working with the restricted version of the modeling problem, where the training data is generated from a numerical PDE solve. A benefit of this approach is that the numerical solution to the PDE given in Eq. \eqref{data_generating} itself defines a coefficient map, and can thus be compared and contrasted with the parameterized coefficient map of the data-driven model. In other words, the true system response and the model of that response have the same mathematical form, where coefficients on a fixed basis are determined as a function of the discretized forcing.

\paragraph{} Having laid out the true system response, we now define the measurement process. Introducing the sensor positions $\{ \mbf y_k\}_{k=1}^{\mathcal S}$ and a measurement matrix $G_{ij} = h_j(\mbf y_i)$, the measurement data is generated with

\begin{equation}\label{measurement}
    \mbf v_i = \mbf G\mbf{\tilde U}(\mbf F_i) + \bs \xi_i, \quad \bs \xi_i \overset{\text{i.i.d.}}{\sim} \Xi(\mbf 0 , \bs \Sigma),
\end{equation}

\noindent where $\bs \xi_i$ is a noise vector whose distribution $\Xi$ is an arbitrary zero mean distribution with covariance $\bs \Sigma$, and $\mbf F_i$ is the discrete representation of the $i$-th input forcing field. The measurement matrix $\mbf G$ simply picks out the value of the state field at the sensor locations. When the solution coefficients $\mbf{\tilde U}$ are determined through the nonlinear solve of Eq. \eqref{data_generating}, we call Eq. \eqref{measurement} the ``data generating process,'' as we conceptualize this as a mathematical model of both the system's response and the measurement of that response.

\section{Training and testing the model}

\begin{figure}[hbt!]
\centering
\includegraphics[width=0.99\textwidth]{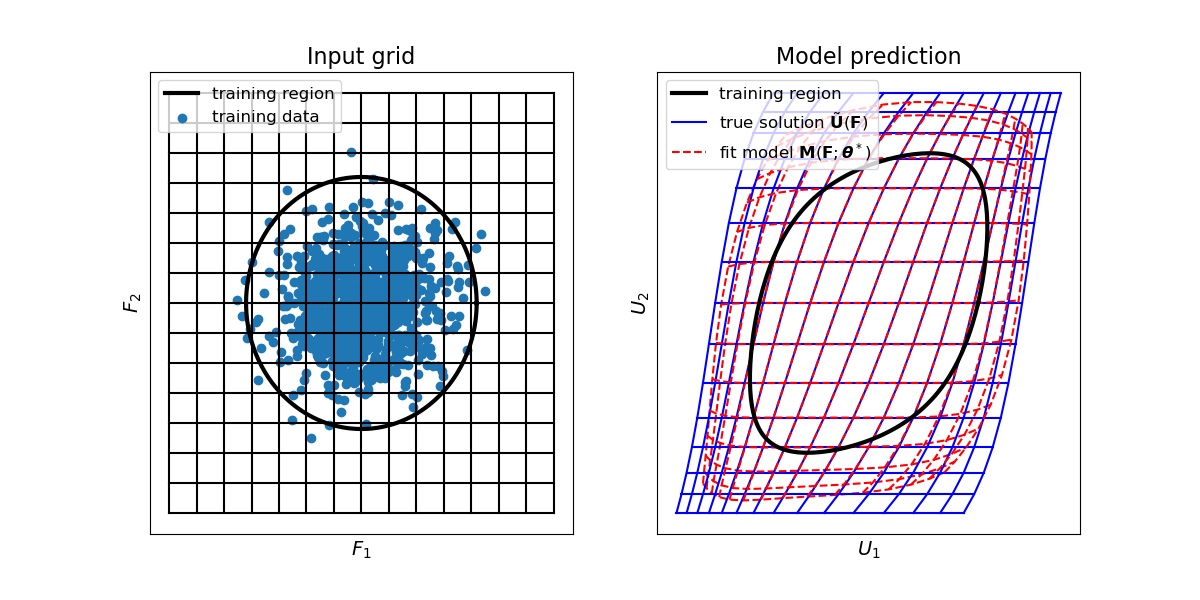}
\caption{The training region is defined to contain a specified fraction of the training inputs, which are drawn from a multivariate normal distribution. In this case, the learned model agrees with the data generating process in the training region, but not when extrapolating.}
\label{generalization}
\end{figure}

\begin{figure}[hbt!]
\centering
\includegraphics[width=0.99\textwidth]{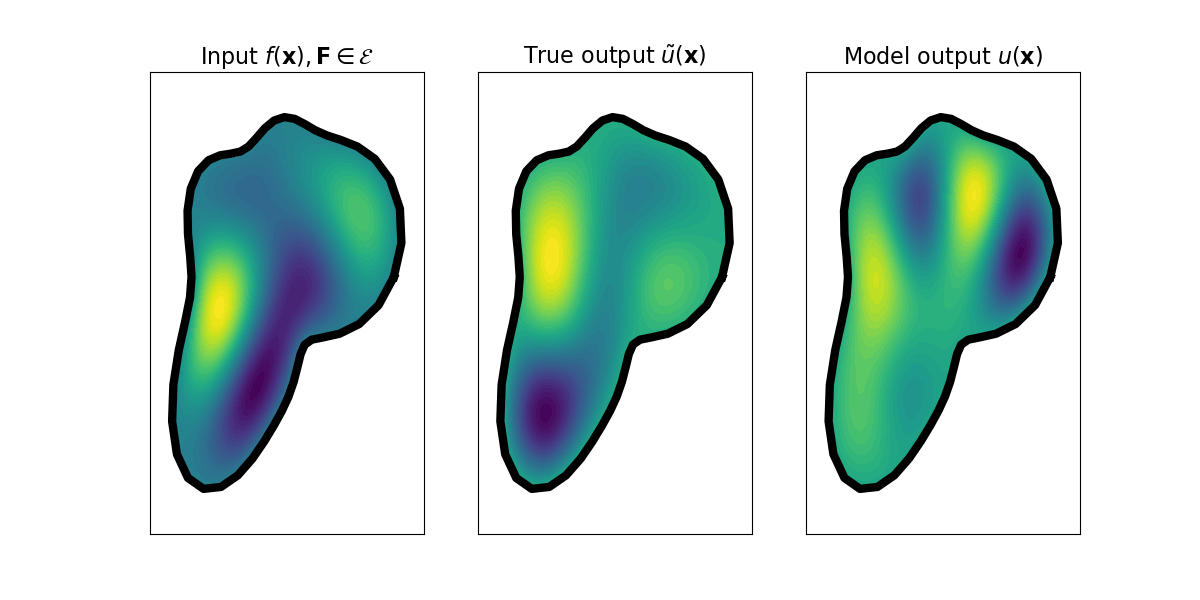}
\caption{A data-driven model may fail to predict the true state field when extrapolating.}
\label{generalization_spatial}
\end{figure}

\paragraph{} At this point, we have used the Allen-Cahn equation to give a specific example of a data generating process, and have specified the general mathematical form of the parameterized model. Still, we remain intentionally vague about the exact mathematical details of the parameterized coefficient map $\mbf M(\mbf F; \bs \theta)$, as this will be the focus of subsequent sections. Now, we can train the model---meaning we can estimate the parameters that define it---by solving an optimization problem for the parameters $\bs \theta$:

\begin{equation*}\label{training}
    \bs \theta^* = \underset{\bs \theta}{\text{argmin }} \frac{1}{2}\sum_{i=1}^N \lVert \mbf G \mbf M(\mbf F_i ; \bs \theta ) - \mbf v_i\rVert^2 + \mathcal R(\bs \theta),
\end{equation*}

\noindent where $\mathcal R(\bs \theta)$ is some regularization on the parameters (sparsity \cite{shea_sindy-bvp_2021}, small norm \cite{zhang_understanding_2017}, etc.), though we note that regularization need not appear. With the discrete representation of the forcing, the training data is $\mathcal D =\{ \mbf F_i , \mbf v_i \}_{i=1}^N$. We will assume that the $\mbf F_i$ come from a convex region, meaning that a line connecting two sets of forcing coefficients lies inside what we call the ``training region.'' For simplicity, we take the training inputs to have been drawn from a multivariate normal distribution with circular covariance, e.g., $\mbf F \sim \mathcal N( \mbf 0 , \sigma^2 \mbf I)$. In this case, we define the training region to be $\mathcal I =\Big \{ \mbf F : \sqrt{\sum_{i=1}^{\mathcal B} F_i^2} \leq 3 \sigma \Big \}$. In general, we think of the training region as the smallest convex space that contains some user-specified high percentage of the training inputs. We note that it is not necessary to obtain training inputs as samples from a normal distribution, but we do require that the $\mbf F_i$ form a convex training region. When the trained model $\mbf M(\mbf F; \bs \theta^*)$ is queried in the training region ($\mbf F \in \mathcal I$), we say that the model is ``interpolating.'' Calling the complement of the training region $\mathcal E = \mathcal I^c$, we say that the model is ``extrapolating'' when $\mbf F \in \mathcal E$. To illustrate this, we take the dimension of the discretization of the input and output fields to be $\mathcal B=2$, meaning that model is  a function $\mbf M: \mathbb R^2 \rightarrow \mathbb R^2$. Borrowing an idea from solid mechanics, we visualize this low-dimensional model with its deformation of a rectangular grid. In particular, we construct a rectangular array of points in the input $\mbf F$ space, and connect them with grid lines. We then pass each of these points through the model, and again connect neighboring points with grid lines, which are now tilted and scaled by the ``deformation'' of the model. The deformed grid provides visual insight into the map defined by the model, and multiple grids can be overlaid to compare different models. See Figure \ref{generalization} for a visualization of example training samples, the training region, the rectangular input grid, and the image of the training region and the grid under the deformation defined by the model. Without particular reasons to do so, we have no reason expect the data-driven model to ``generalize,'' meaning that it does not agree with the true data generating process outside the training region. Whereas the grid deformation shows interpolation and extrapolation over a range of inputs, Figure \ref{generalization_spatial} provides the spatial perspective on the failure of generalization, in which the forcing and state fields are reconstructed with their coefficients. In general, when the model is queried in the extrapolation regime, the true and model output fields do not agree with each other. Connecting back to the philosophical literature, a model which merely summarizes training data, but does not make accurate predictions outside the training region, may be called phenomenological \cite{mcmullin_what_1968, termine_machine_2026}.

\paragraph{} We claim that it is intuitively clear that, in the case of Figure \ref{generalization_spatial}, the data-driven model does not generalize. However, it is too strict a requirement to define generalization as perfect agreement between the true system response and the model when extrapolating. We thus define the condition for generalization as

\begin{equation}\label{generalization_eq}
     \underbrace{\Big| \mathbb E\Big[ \mbf G \mbf{\tilde U}(\mbf F) + \bs \xi - \mbf G \mbf M(\mbf F;\bs \theta^*)\Big] \Big |}_{\text{expected discrepancy magnitude}} = \Big| \mbf G\Big( \mbf{\tilde U(\mbf F) - \mbf M(\mbf F;\bs \theta^*)}\Big)\Big|  \leq \mathcal T, \quad \mbf F \in \mathcal E^*,
\end{equation}

\noindent where the norm here is $|\mbf x|=\sqrt{\sum_ix_i^2}$ and we use that the measurement noise has zero expectation. The quantity $\mathcal T$ is some user-specified threshold that controls the stringency of the demand on accuracy in the extrapolation region and $\mathcal E^* \subset \mathcal E$ is a subset of the extrapolation region of interest to the modeler. We restrict ourselves to the region $\mathcal{E}^*$ because it is often the case that the model predictions increase with $|\mbf F|$, such that discrepancies between the data-driven model and the true coefficients can grow arbitrarily large, even when the model is reasonably accurate in some finite region. 

\begin{figure}[hbt!]
\centering
\includegraphics[width=0.9\textwidth]{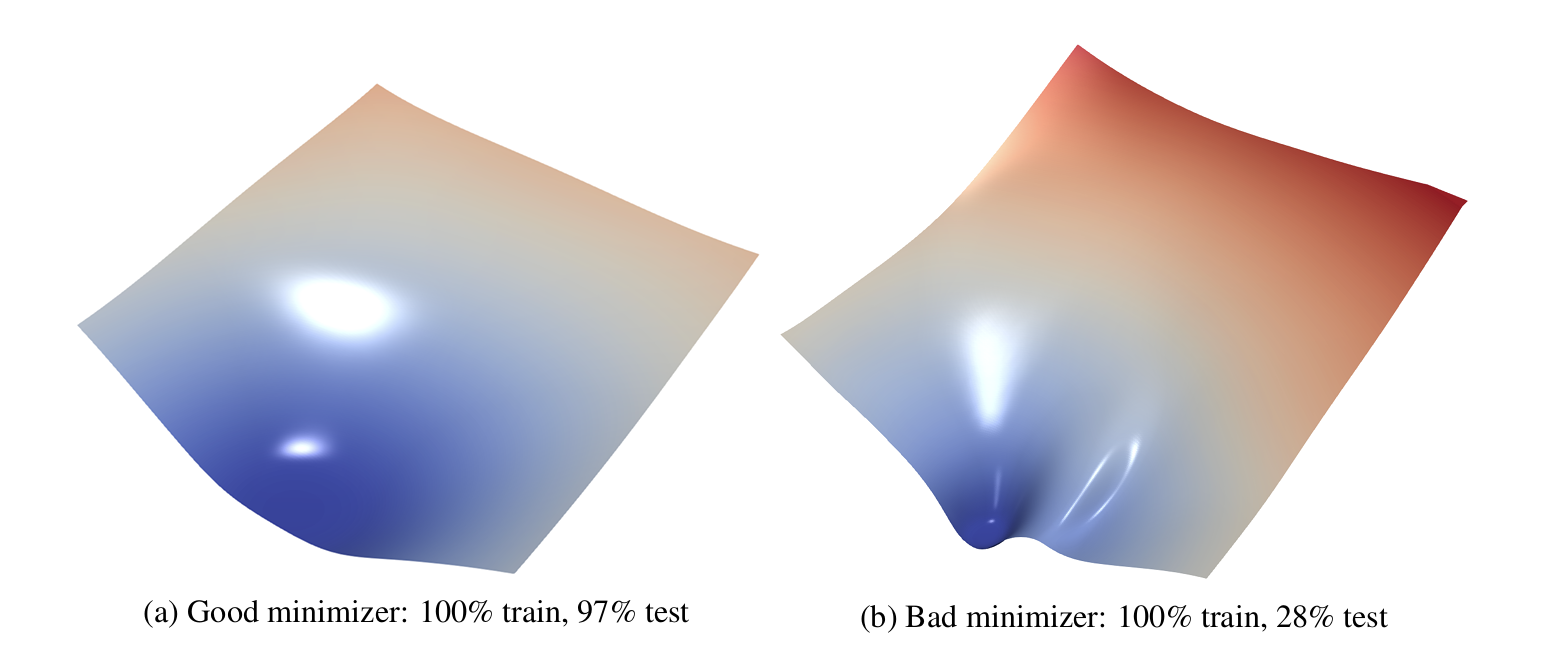}
\caption{When discussing generalization of a data-driven, authors in mainstream machine learning (as opposed to scientific machine learning) refer to the performance of the model on unseen inputs, not to inputs which are explicitly ``outside'' of the training data. One approach to study generalization in this sense is with the loss surface of the training objective. Image adapted from \cite{huang_understanding_2020}.}
\label{flat_minima}
\end{figure}

\begin{figure}[hbt!]
\centering
\includegraphics[width=0.99\textwidth]{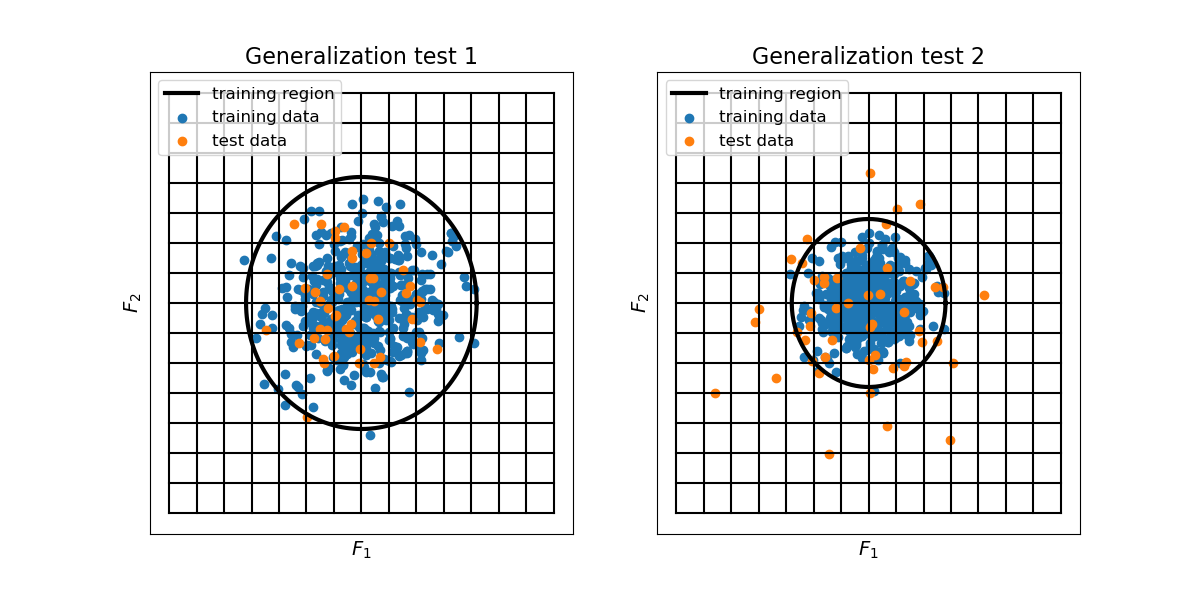}
\caption{In machine learning-based classification problems, the generalization error of the model is determined by its performance on unseen inputs which are statistically similar to the training data (test 1). In science, we take generalization to mean that the model performs well even on inputs which are not statistically similar to the training data (test 2).}
\label{generalization_test}
\end{figure}

\paragraph{} We note that our notion of generalization is quite different than that of the mainstream machine learning literature. Here, a common task is to build a model to determine which among a finite set of classes a piece of data belongs to. Famously, neural networks can be trained to take in images and predict the type of animal depicted in the image. In this context, the generalization has to do with the performance of the model on images in the ``test set,'' a fraction of the data intentionally excluded from the process of training the model, which is used to assess the model's capabilities on unseen tasks. We emphasize that the test data for classification problems is assumed to be statistically similar to the data used in training. For example, when the authors in \cite{huang_understanding_2020} discuss how different local minima of the loss surface of the training objective have different generalization properties, they refer to classification performance on test data that is statistically similar to the training data.\footnote{This paper contributes to an ongoing debate in the machine learning literature about how the ``flatness'' of the region of the loss surface surrounding the solution found by the optimizer influences the test performance of the model. In particular, the authors argue that flat minima lead to simpler decision boundaries between classes, which translates to better performance on unseen inputs.} See Figure \ref{flat_minima} for an example of their loss surface analysis, which we take as representative of discussions of generalization in the classification literature. Similarly, the well-known ``double descent'' phenomenon states that highly overparameterized data-driven models generalize better than models with fewer parameters, contrary to the prevailing intuition that they should overfit \cite{belkin_reconciling_2019}. Again, generalization refers to unseen inputs, rather than inputs which are outside the convex hull of the training data. In scientific modeling, we hold that our more demanding definition of generalization is appropriate, as scientists are accustomed to models which can be used in any situation which respects the assumptions of the model (small strains, non-relativistic velocities, etc.), without referring to the details of the training process. An engineer doing structural analysis does not need to know the exact deformation of a test coupon used to calibrate the stiffness (Young's Modulus) of a material, as the equations of elasticity are taken to be so general that these details do no matter. See Figure \ref{generalization_test} for a schematic of the classic machine learning definition of generalization, and our stricter definition of generalization for scientific problems. 

\paragraph{} We remark that our definition of generalization, and thus our analysis of the conditions under which it is possible, also conflict with notions of model class and generalization put forward by statistical learning theory. For example, as stated in \cite{luxburg_statistical_2011}, statistical learning theory asks ``what learning tasks can be performed by computers?'', ``what kind of assumptions do we have to make such that machine learning can be successful?'', and ``what are the key properties that a learning algorithm needs to satisfy in order to be successful?'' Once again, the notion of success here has to do with accurately interpolating the training data, rather than generalizing (in our sense) outside of the training data. This is seen in the assumption that the training data is drawn from the same statistical distribution as downstream test data. Thus, the fundamental questions of statistical learning theory differ from ours in this work. Furthermore, we question whether foundational results of statistical learning theory apply to differential equation-based models of physical systems. For example, the bias-variance tradeoff states that fits given by more complex/expressive models are more sensitive to noise than simpler models. It is not clear that this conclusion applies to models where parameters represent physical processes in the system of interest, as the behavior of the model is often very circumscribed even when the number of candidate processes is large. To reiterate, we maintain that questions about data-driven models for physical phenomena need not overlap with questions about data-driven models writ large.

\paragraph{} An interesting, idealized test case for these ideas is when there is zero measurement noise, and the measurement matrix is invertible, meaning that there must be as many measurements taken as there are basis functions $\mathcal B$. In this case, we can omit the measurement matrix $\mbf G$, as it has no effect on optimization problem for the model parameters. Training then becomes

\begin{equation}\label{idealized_training}
    \bs \theta^* = \underset{\bs \theta}{\text{argmin }} \frac{1}{2}\sum_{i=1}^N \lVert  \mbf M(\mbf F_i ; \bs \theta ) - \mbf{\tilde U}(\mbf F_i)\rVert^2 + \mathcal R(\bs \theta).
\end{equation}

\begin{figure}[hbt!]
\centering
\includegraphics[width=0.99\textwidth]{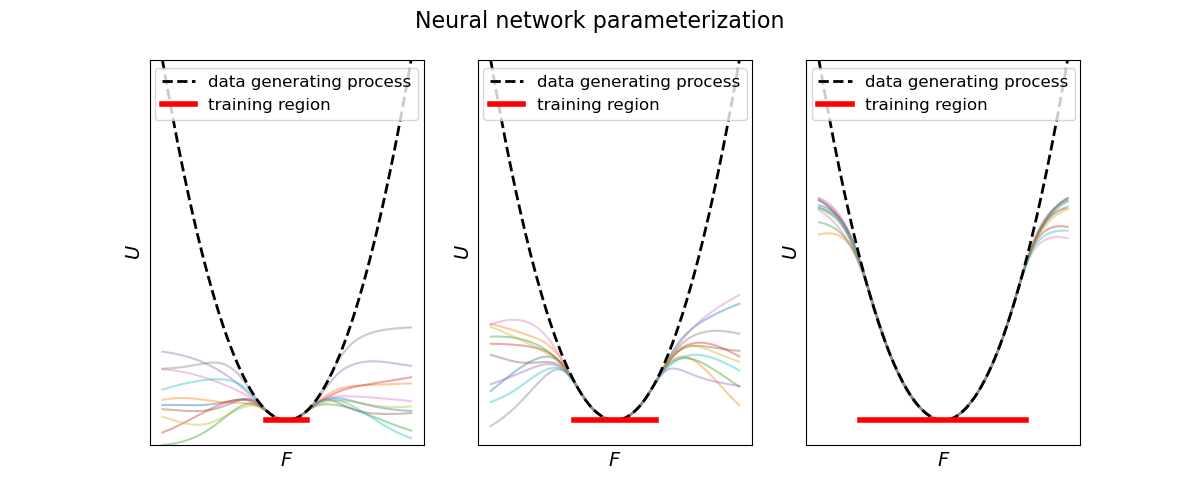}
\caption{For each training region size, $10$ neural networks with random initialization are trained to match data taken noiselessly from the data generating process. Regardless of the size of the training region, the networks tend to predict different extrapolation behavior, even while agreeing in the training region. This suggests that the model is non-identifiable given the training data, which we take to rule out the possibility of generalization. }
\label{identifiability}
\end{figure}

In this idealized setting, we now ask: when is it possible for the parameterized model to generalize based on a finite set of $N$ experimental trials? First, we claim that a necessary but not sufficient condition is that the parameterized model is \textit{identifiable} given the training data \cite{wieland_structural_2021, maclaren_what_2020}. In simple terms, this means that there is a unique solution $\bs \theta^*$ to the optimization problem used to train the model in Eq. \eqref{idealized_training}. It may be the case that multiple parameter settings explain the training data equally well, but predict different behavior when extrapolating.\footnote{In some cases, different solutions to a non-identifiable inverse problem may correspond to equivalent models. For example, $U=\theta_1 \theta_2 F$ is non-identifiable, but different parameter settings build the exact same model. We assume that the non-identifiable set of parameters corresponds to different models, and thus different extrapolation behavior.} In this case, the calibrated model cannot be taken to generalize reliably, as extrapolatory predictions vary depending on which solution to the optimization problem is found. The non-identifiability of one-dimensional neural networks models is shown in Figure \ref{identifiability} for three different sized training regions. Depending on how the networks are initialized, they obtain the same fits of the noiseless data generating process in the training region, but make different predictions when extrapolating. If the data is not sufficient to uniquely identify the parameters of the model, then there is no reason to expect generalization, as the generalization behavior of the model is not even uniquely defined.

\paragraph{} To see why identifiability is not sufficient for generalization, we note that identifiability simply means that there is a unique solution to the model calibration problem, not that the data generating process can be globally recovered. For example, parameters of a model are identifiable when they are the unique best approximation of the training data within the prescribed model class. As shown in Figure \ref{regression}, an example of identifiability but not generalization arises when performing linear regression on data generated from a quadratic function. Thus, seeing that identifiability is insufficient for model generalization, we make the following claim about data-driven models:

\begin{figure}[hbt!]
\centering
\includegraphics[width=0.7\textwidth]{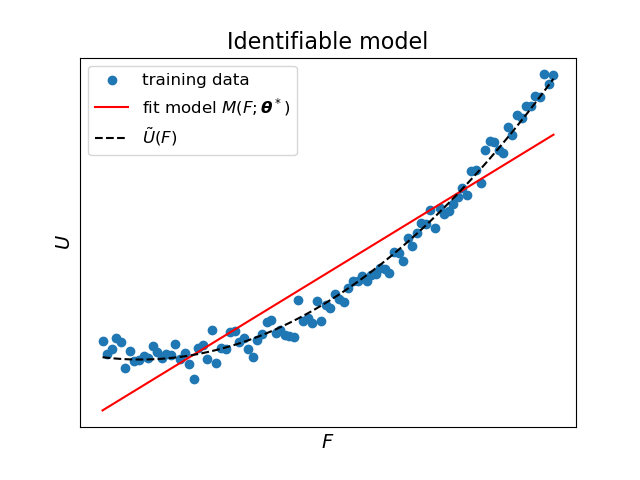}
\caption{The slope and intercept of a linear model are identifiable with sufficient training data, regardless of whether this model class matches the data generating process. In this case, the linear model will not predict the quadratic behavior of data generating process, either inside or outside of the training region.}
\label{regression}
\end{figure}

\paragraph{Claim} Data-driven models only generalize when the model class of the parameterized model is equivalent to the model class of the data generating process and the parameters are identifiable.

By model class, we mean that the mathematical form of the potentially nonlinear parameterized function mapping the discretized forcing to the solution coefficients. In other words, for the model class of the data-driven model and the true system response to be equivalent, it is necessary that $\mbf M(\mbf F;\bs \theta) = \mbf{\tilde U}(\mbf F)$ for some parameter setting $\bs \theta$ (this is called ``structural equivalence'' in \cite{naser_what_2026}). Note that this is a much stronger condition than Eq. \eqref{idealized_training}, which states only that the model optimally approximates the system response in the training region. We claim that for generalization to occur, it is necessary that the modeler somehow anticipates the mathematical form of the data generating process. For example, returning to the Allen-Cahn equation, we claim that all models capable of generalizing have the following form:

\begin{equation*}
    \mbf M(\mbf F ; \bs \theta) = \underset{\mbf{V}}{\text{solve}}\Big( -\theta_1 \mbf K \mbf{V} +\theta_2 \mbf A \mbf V - \theta_2 \mbf R \vdotop  ( \mbf{V} \otimes \mbf{V} \otimes \mbf{V}) + \mbf N(\mbf V ; \theta_3,\theta_4,\dots) + \mbf F = \mbf 0\Big),
\end{equation*}

\noindent where $\mbf N(\mbf V; \theta_3,\theta_4,\dots)$ is an arbitrary nonlinear function of the state involving all parameters except those on the terms arising from the discretized Allen-Cahn PDE. The true solution, given in Eq. \eqref{data_generating}, can thus be recovered with an appropriate choice of the model parameters. We can appreciate why the definition of generalization is restricted to a subset of extrapolation region $\mathcal E^*$, as even when the parameters are such that $\mbf N=\mbf 0$, any non-zero error $|\kappa-\theta_1| >0$ and $|\sigma-\theta_2| >0$ can cause arbitrary large errors in the model prediction as $\mbf F \rightarrow \infty$. Thus, we specify a finite extrapolation region of interest, such that imperfect recovery of coefficients does not prevent us from considering a data-driven model to be generalizable. If the parameterized coefficient map is not a nonlinear solve, we take this to guarantee that the model cannot generalize, owing to the mismatch of model classes. Of course, this does not guarantee that all coefficient maps defined by nonlinear solves are capable of generalizing.

\paragraph{} It is not clear how to definitively prove our claim that generalization is only possible when the parameterized model matches the model class of the data generating process, but we remark that the modeler would have to get \textit{extraordinarily lucky} if the wrong model class happened to make the correct predictions in the extrapolation region. This is because, by definition, the parameterized model has not been trained to make predictions here. In essence, we argue that generalization behavior is built into a model by choosing the correct model class, which is often possible with the help of prior knowledge. Generalization is thus hard-coded, it cannot be learned, no matter how much data a model is trained on. Extensive experience with physical systems indicates that parsimonious and simple PDEs are a powerful model class, which very often generalize outside of the training region they are calibrated on. This generality of PDEs is what allows modelers to talk of learning ``material properties'' of a system when performing calibration, as the PDE is so general that the model parameters not only generalize with respect to input forcings, but also to different geometries, and settings where other physical effects are present.\footnote{For example, the elastic stiffness of a material can be calibrated on an experimental coupon, and then used to predict fracture of a body with different geometry \cite{bourdin_numerical_2000}. This is even when the initial model calibration problem did not include the physics of fracture.} In the philosophical literature, many authors would argue that a data-driven models whose model class matches the data generating process, (and thus has the capacity to generalize), is capable of discovering mechanisms \cite{craver_ontic_2014, glennan_rethinking_2002, bechtel_explanation_2005}. This is because the model parameters actually have physical meaning, describing the ``entities'' and ``activities'' that comprise system components. Note that the emphasis in the philosophical literature on a system being built from parts or modules maps nicely on to the example of the PDE model class, as these models have an explicit commitment to describing local interactions at ``material points.'' Many of the philosophers of science interested in mechanism have backgrounds in biology and neuroscience, so the mechanisms they have in mind are of finite size (cell interactions, neurons, etc.). The only meaningful difference we see between biological mechanisms of this sort and the mechanisms defined by PDE models is that the entities and activities of the PDE system exist at infinitesimal length scales. However, the idea that the behavior of the system is produced by the ``interaction of a number of parts, where the interactions between parts can be characterized by direct, invariant, change-relating generalizations'' remains intact \cite{glennan_rethinking_2002}. This is precisely the idea of applying conservation laws to regions of a body encountered in continuum mechanics. For further exposition of a mechanistic decomposition of a mass transport system, see Appendix \ref{sec: conservation}. With these ideas in mind, we note that the difference between a phenomenological model and a mechanistic model is nothing more than the model class of the coefficient map $\mbf M(\mbf F; \bs \theta)$. Going forward, we take parameterized coefficients maps with the correct model class to discover mechanisms when the parameters are identifiable. Returning to our discussion on inferentialism and interventionism, we argue that for scientific PDE models, a generalizable model both allows for the modeler to robustly draw inference---even inferences outside the training data---and to have causal understanding of the system of interest, as we take to be the entity/activity perspective on the system of interest to describe causal processes \cite{dowe_wesley_1992}. As such, our account shows that the ideas of generalization of data-driven models, mechanisms, and causality are all intimately intertwined.

\paragraph{} We have argued that generalization, in the sense we have defined it, is only possible when the class of the parameterized model matches the model class of the data generating process. By matching model classes, we mean that there is some parameter setting for which the parameterized model exactly recovers the mathematical form of the data generating process. In the case of scientific data generated from PDEs, this means that, at the very least, the parameterized model must be defined implicitly through a nonlinear solve. Now, we turn to data-driven modeling strategies from the scientific machine learning literature to investigate the assumptions they make about model classes, and thus their prospects for generalization.

%------------------------------------------------------------------------------

\section{Data-driven modeling strategies}

\paragraph{} Traditionally, PDEs formulated from deep domain knowledge with a small number of empirical parameters have dominated the practice of scientific modeling. In recent years, a number of other data-driven modeling strategies have emerged. In this section, we review four distinct data-driven modeling strategies, cast them in the form of Eq. \eqref{model}, and assess the model class of the coefficient map to understand if the model is capable of generalizing. Throughout, we assume that the true data generating process is the discretized solution to a parsimonious and simple PDE.

\subsection{Inverse problems}

\paragraph{} Inverse problems are the most traditional technique for building data-driven models. We reiterate that while many PDEs in science are derived from first principles---e.g., conservation of mass, momentum, and energy---they still depend on empirical parameters, usually material properties (see Appendix \ref{sec: conservation}). As such, we treat PDEs as one among many types of data-driven model. In particular, they are a form of data-driven model informed extensively by prior knowledge, and consequently have a circumscribed mathematical form. It is often the case that there are only a few parameters building the PDE model in the inverse problem, and it is often assumed that this parameterized model exactly mirrors the data generating process. One can appreciate this fact by noting that specialized techniques are required to handle PDE inverse problems in the case of ``mis-specified models,'' meaning that the model class of the PDE and data generating process are different \cite{sargsyan_statistical_2015, rowan_extending_2025}. Continuing with the example of the Allen-Cahn equation, we assume that the parameterized model is

\begin{equation*}
\begin{aligned}
    \theta_1 \nabla^2 u(\mbf x) + \theta_2 u(\mbf x)( 1 - u(\mbf x)^2) + f(\mbf x) = 0, \quad \mbf x \in \Omega, \\
    u(\mbf x) = 0, \quad \mbf x \in \partial \Omega,
\end{aligned}
\end{equation*}

\noindent where $\bs \theta=[\theta_1,\theta_2]^T$ are the parameters of the model. Discretizing the input and output field, and following Appendix A, the solution coefficients are governed by a solution to the weak form system of equations:

\begin{equation*}
    \mbf {U}^{\text{INV}}(\mbf F) = \mbf M^{\text{INV}}(\mbf F; \bs \theta) = \underset{\mbf{V}}{\text{solve}}\Big( -\theta_1 \mbf K \mbf{V} +\theta_2 \mbf A \mbf V - \theta_2 \mbf R \vdotop  ( \mbf{V} \otimes \mbf{V} \otimes \mbf{V}) + \mbf F = \mbf 0\Big).
\end{equation*}

\begin{figure}[hbt!]
\centering
\includegraphics[width=0.99\textwidth]{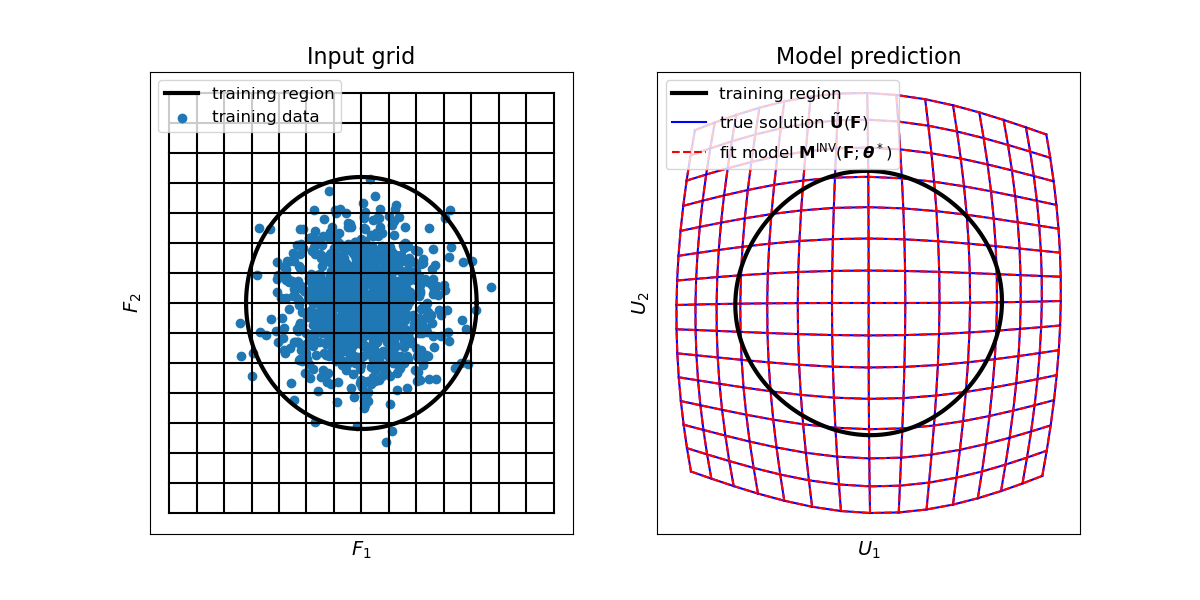}
\caption{Because the model class of the inverse problem is equivalent to the data generating process, the fit model generalizes outside the training data. According to us, the recovered PDE then leads to a mechanistic understanding of the processes giving rise to the training data.}
\label{AC_INV}
\end{figure}

We note that when the model is correctly specified in an inverse problem, the parameterized model and data generating process have the same model class. This means that if there is sufficient data for the parameters to be identifiable, the model built by an inverse problem generalizes, and can thus be said to discover the mechanism underlying the training data. This is reflected in scientific practices, as models built from inverse problems are often used to make predictions without concern for whether the prediction is interpolating (making predictions inside the training region $\mathcal I$) or extrapolating (generalizing, predictions in $\mathcal E$). The generalization of the learned model is illustrated with a two-dimensional coefficient map in Figure \ref{AC_INV}.

\subsection{SINDy}

\paragraph{} Introduced in 2016, a method called the ``sparse identification of nonlinear dynamics'' (SINDy) extends inverse problem to settings where the exact model class is not known a priori \cite{brunton_discovering_2016}. While introduced originally for ordinary differential equations, the basic methodology was soon extended to PDEs \cite{rudy_data-driven_2016}. With SINDy, the governing equation is represented as a linear combination of a user-defined set of ``library'' terms:

\begin{equation*}
\begin{aligned}
    \begin{bmatrix}
        \theta_1 \\ \theta_2 \\ \theta_3 \\ \theta_4 \\ \theta_5 \\ \vdots
    \end{bmatrix} \cdot \underbrace{\begin{bmatrix}
        u \\ u^3 \\ \partial u / \partial x_1 \\ \partial u / \partial x_2 \\ \nabla^2 u  \\ \vdots
    \end{bmatrix}}_{\text{library of candidate terms}} + f(\mbf x) = 0, \quad \mbf x \in \Omega, \\
    u(\mbf x) = 0, \quad \mbf x \in \partial \Omega.
\end{aligned}
\end{equation*}

\begin{figure}[hbt!]
\centering
\includegraphics[width=0.99\textwidth]{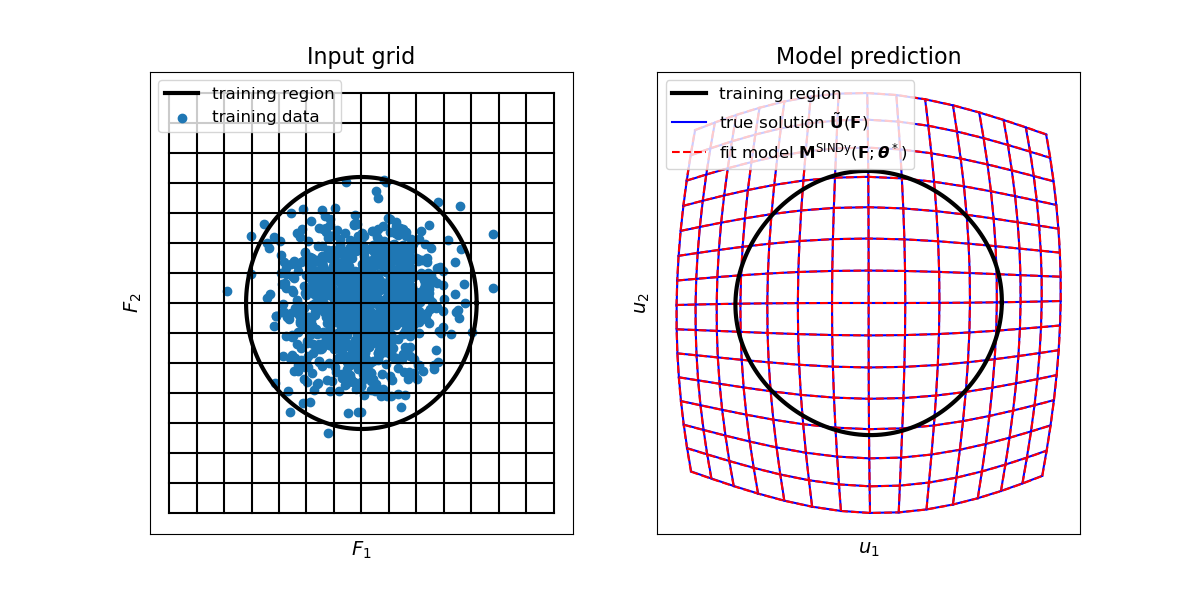}
\caption{While the model class of the SINDy parameterization is more flexible than the inverse problem, the discretized Allen-Cahn model can still be exactly recovered with the appropriate choice of library. As such, the learned model can generalize and discover mechanisms.}
\label{AC_SINDy}
\end{figure}

Again discretizing the input and output fields in a basis, the solution of the weak form of the PDE represented with the SINDy library is given as

\begin{equation*}
    \mbf U^{\text{SINDy}}(\mbf F) = \mbf M^{\text{SINDy}}(\mbf F;\bs\theta) = \underset{\mbf V}{\text{solve}} \Big(  \sum_{i=1}^{\mathcal L}  \theta_i \mbf L_i(\mbf V) + \mbf F = \mbf 0\Big),
\end{equation*}

\noindent where $\mathcal L$ is the number of library terms and $\mbf L_i(\mbf V)$ is the discretized form of the $i$-th library term. Unlike an inverse problem, many of the optimal model parameters $\bs \theta^*$ are expected to be zero with the SINDy parameterization, which is enforced with a sparsity regularization on the coefficient vector \cite{donoho_compressed_2006}. That being said, SINDy assumes that the modeler has sufficient knowledge of the system under study for the true model to be obtained as some linear combination of model terms. Given the larger number of library terms, the question of identifiability becomes more complicated in the case of SINDy, but, along with other authors in the literature, we assume that the true model is contained in the SINDy library and becomes identifiable with sufficient training data. As such, data-driven models built with SINDy are capable of generalizing, and thus discovering mechanisms underlying the training data. In the case of the Allen-Cahn equation, and with a SINDy library of $[ u,u^2,u^3,\partial u/\partial x_1 , \partial u / \partial x_2 , u \partial u/\partial x_1 , u \partial u / \partial x_2 ,\partial^2 u / \partial x_1^2 , \partial^2 u / \partial x_2^2]^T$, the recovered coefficients lead to a model that generalizes, as shown in Figure \ref{AC_SINDy}.

\subsection{Neural BVP}

\begin{figure}[hbt!]
\centering
\includegraphics[width=0.99\textwidth]{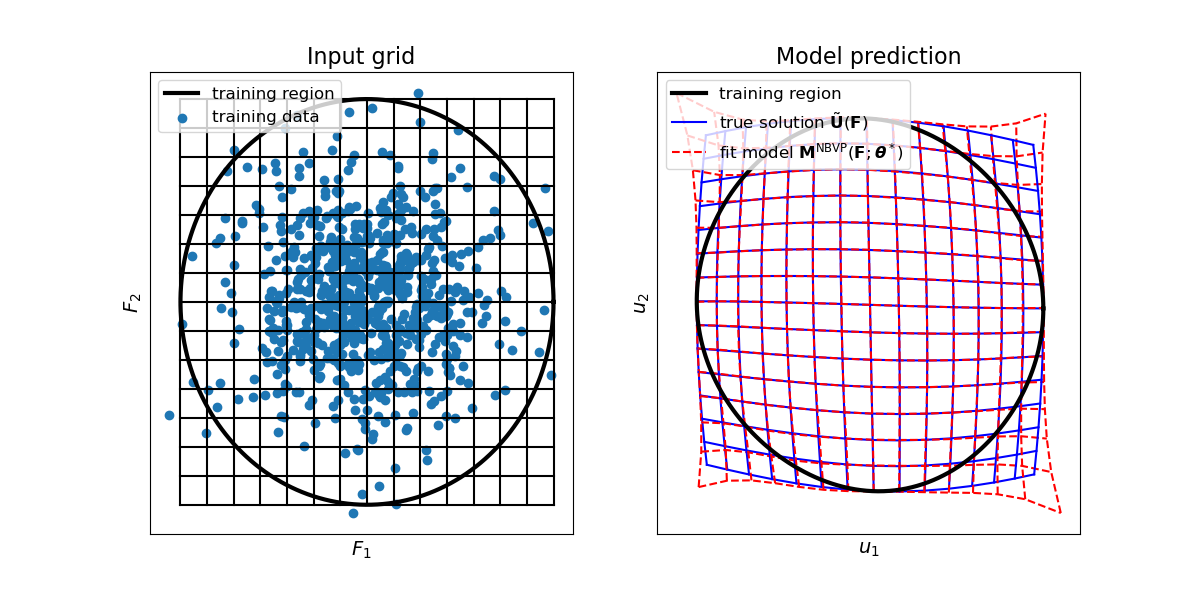}
\caption{Based on our assumption that PDEs governing physical systems are simple, a neural network model of the differential operator is never the correct model class. Though it can fit the training data, it does not generalize, and thus, does not discover mechanisms.}
\label{AC_NBVP}
\end{figure}

\paragraph{} Whereas SINDy relaxes the assumption that all parameters of the PDE appear in the final model, an alternative strategy called ``neural ordinary differential equations'' lifts the requirement that a library of candidate terms containing the true model can even be written down. Named as such because it was introduced in the context of ordinary differential equations, this method represents the differential operator as a neural network taking in the output field and its derivatives \cite{chen_neural_2019}. This method has been studied in a variety of contexts, such as fluid mechanics \cite{wang_learning_2024} and reactor systems \cite{sorourifar_physics-enhanced_2023}. Because we employ this basic idea in the context of boundary value problems (BVPs), we call this method ``Neural BVP.'' In this context, the governing equation is 

\begin{equation*}
\begin{aligned}
    \mathcal N\qty( u ,  \pd{u}{x_1},\pd{u}{x_2},\pdd{u}{x_1}, \pdd{u}{x_2},\dots; \bs \theta) + f(\mbf x) = 0 , \quad \mbf x \in \Omega, \\
    u(\mbf x) = 0 , \quad \mbf x \in \partial \Omega,
\end{aligned}
\end{equation*}

\noindent where $\mathcal N$ is a fully-connected neural network, the mathematical details of which are shown in Appendix \ref{sec: nn}. Note that the differential operator is no longer a linear combination of the parameters. As shown in Appendix \ref{sec: nbvp}, this means that the weak form does not consist of a sum of discretized differential terms with constant coefficients. The governing equation for the solution coefficients is

\begin{equation*}
    \mbf U^{\text{NBVP}}(\mbf F) = \mbf M^{\text{NBVP}}(\mbf F;\bs \theta) = \underset{\mbf V}{\text{solve}}\Big(  \mbf N(\mbf V ; \bs \theta) + \mbf F = \mbf0 \Big),
\end{equation*}

\noindent where $\mbf N(\mbf V ; \bs \theta)$ is the discretized form of the differential operator defined by a fully-connected network with parameters $\bs \theta$.  Recall that, based on past scientific experience, we have claimed that PDEs modeling physical phenomena are ``simple'' in the very minimal sense that there are not compositions of nonlinear functions. If the neural network $\mathcal N$ approximating the differential operator is ``deep,'' meaning that it has multiple layers, there will be compositions of nonlinear functions. This means that though Neural BVPs still represent a PDE model of the data, the representation of the differential operator is a different model class than the data generating process. This means that this type of data-driven model cannot generalize, despite the fact that the model class involves a nonlinear solve. This is even in the rare case that the neural network parameters can be uniquely identified. While the universal approximation theorem of neural networks guarantees that the network can construct a differential equation whose solution approximates the training data with arbitrary accuracy, it does not guarantee that the learned representation generalizes. As such, we think of neural BVPs as an unusual kind of phenomenological model, which, while encoding the structure of a PDE solve, must be seen as merely summarizing the training data. Figure \ref{AC_NBVP} shows that solutions to the learned differential equation quickly depart from the true solution outside the training region, suggesting that the restriction to the PDE model class does not provide useful regularization to the model-building problem.

\subsection{Neural operators}

\paragraph{} With Eq. \eqref{model}, we claimed that data-driven models can be understood as discretizing the output field in a fixed basis. We noted that in the context of PDE models, this is a common technique underlying widely-used numerical solution strategies such as finite element and spectral methods. The previous three model types all relied on building PDE models of the data, which we then solved in the context of a fixed basis. In contrast to these past three strategies, neural operators forego building a PDE model of the data, and learn an input-output map directly from the data. There are a wide variety of neural operator architectures, and it is not clear at first glance that the fixed basis representation of the output field is always applicable. Thus, before proceeding, we must first review some common neural operators and argue that these models learn a map between the input and coefficients of the output with respect to a fixed basis.

\paragraph{} Introduced in 2019, the Deep Operator Network (DeepONet) was the first popular neural operator, expanding the solution with a trainable set of basis functions, and learning from the data the mapping between the input and the coefficients on this basis \cite{lu_deeponet_2021}. The vanilla DeepONet does not respect the basis expansion assumption, because the basis functions are trainable. However, we remark that one study replaced the so-called ``trunk net'' of the DeepONet, where the basis functions are typically learned, with fixed basis functions obtained with a singular value decomposition of the data. The authors found that the DeepONet with the fixed basis was more accurate and efficient to train the vanilla DeepONet with learnable basis functions \cite{lu_comprehensive_2022}. Thus, we propose that it is reasonable to interpret this architecture in the fixed basis framework. 

\paragraph{} A follow-up neural operator architecture is the Fourier Neural Operator (FNO), which repeatedly convolves the input forcing with trainable integral kernels and passes them through nonlinear activation functions \cite{li_fourier_2021}. The method gets its name because the discrete Fourier transform is used to expedite the computation of convolution integrals. At first glance, this method appears to challenge the fixed basis assumption. However, at the final layer of the FNO, the output field is written as an inverse discrete Fourier transform of a set of coefficients which are a parameterized function of the Fourier coefficients of the input field. This is exactly the discretized representation of the model we wrote in Eq. \eqref{model}, with the proviso that the basis functions are the complex Fourier basis. Along similar lines, by taking the Laplace transform of the forcing and solution, the Laplace Neural Operator (LNO) thus also commits to a complex basis, learning the mapping in frequency space \cite{cao_lno_2023}. 

\paragraph{} A number of other neural operator architectures make the fixed basis representation more explicit. In the Laplacian Eigenfunction Neural Operator (LENO), the input and output fields are discretized in a basis of Laplace eigenfunctions (as in Figure \ref{eigenfunctions}), and the model is taken to be a fully-connected neural network relating the input to output coefficients \cite{hao_laplacian_2025}. Similarly, the so-called ``PCA-net'' builds a basis from principal components of the training data, and learns a neural network between PCA coefficients of the input and output fields \cite{lanthaler_operator_2023}. The basis-to-basis framework also builds basis functions from the training data, and learns a discretized coefficient map \cite{ingebrand_basis--basis_2024}. Beyond Laplacian eigenfunctions, neural operators based on other user-specified basis functions have been studied, for example finite element bases \cite{zhang_finite_2025} and wavelet bases \cite{tripura_wavelet_2023}. All of these examples unambiguously follow the model form introduced in Eq. \eqref{model}.

\begin{figure}[hbt!]
\centering
\includegraphics[width=0.99\textwidth]{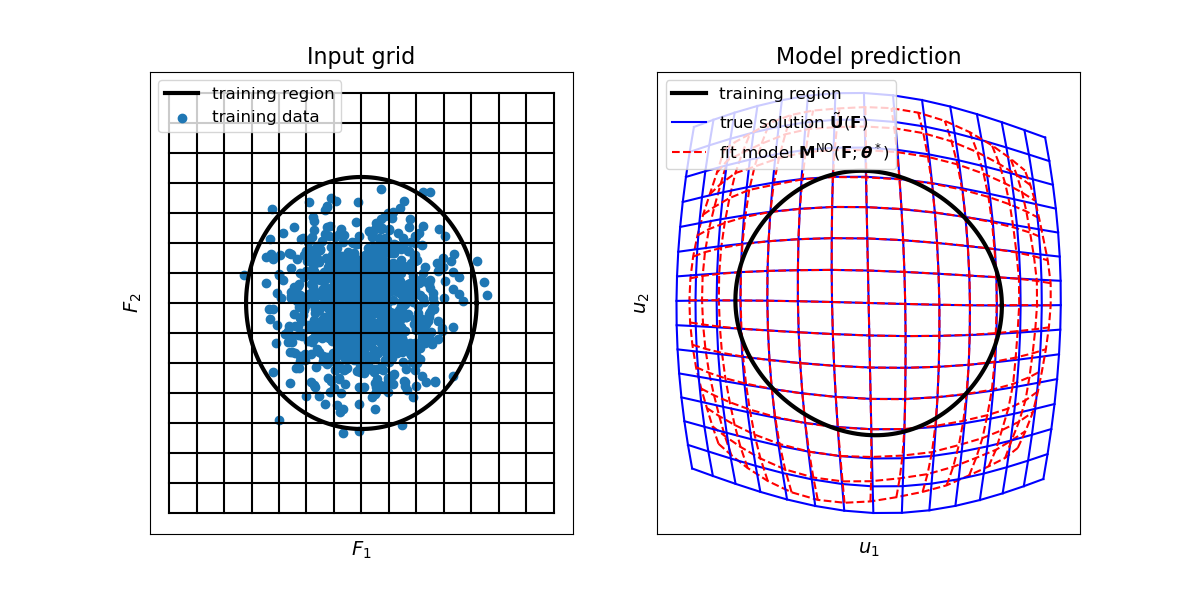}
\caption{Because the model class of PDEs involves nonlinear solves, the neural network representation of the coefficient map is not the correct model class, and thus does not generalize. It is interesting to note that, in our Allen-Cahn example, the neural operator generalizes better than the neural boundary value problem, even though the latter makes predictions through a nonlinear solve.}
\label{AC_NO}
\end{figure}

\paragraph{} Just as the fixed basis assumption does not apply unambiguously to all numerical methods, it may not apply to all neural operators. However, we claim that it is a reasonable mental model for how a majority of operator learning strategies discretize the model-building problem. Unlike the previous three modeling strategies, the coefficient map for neural operators is no longer defined through a nonlinear solve, as no governing differential equation is formulated. As such, we simply write

\begin{equation*}
    \mbf U^{\text{NO}}(\mbf F) = \mbf M^{\text{NO}}(\mbf F ; \bs \theta),
\end{equation*}

\noindent where $\mbf M^{\text{NO}}(\mbf F; \bs \theta)$ is understood to be a fully-connected deep neural network parameterized by weights and biases $\bs \theta$. Now, the true data generating process is a nonlinear solve, and model is a forward pass through a neural network. Thus, even if the neural network parameters are identifiable, the model classes are different, and the neural operator is not capable of generalizing. This is illustrated in Figure \ref{AC_NO}, where the learned coefficient map accurately approximates the true solution in the training region, but then quickly diverges from it when extrapolating. We claim that neural operators should be thought of as phenomenological models, which summarize the training data, and allow for interpolation of it, but do not discover mechanisms of the system and thus do not generalize. The above examples highlight that there is a tension between the flexibility of a data-driven models---its ability to represent a diverse set of input-output relations---and its ability to generalize. This is because making assumptions about the model class of the data, while necessary for generalization, limits the flexibility, or expressiveness, of the parameterized model.

%------------------------------------------------------------------------------

\section{Discussion}

\begin{table}[hbt!]
\renewcommand{\arraystretch}{2.3}
\centering
\caption{Comparing the four data-driven modeling strategies in the case of the Allen-Cahn example BVP. Here, we define the vector of basis functions as $\mbf h(\mbf x) = [h_1(\mbf x), \dots, h_\mathcal B (\mbf x)]^T$.}
\label{tab:summary}
\begin{tabular}{|c|c|c|c|c|}
\hline
 & Model form & Coefficient map & Meaning of parameters & Generalization? \\ \hline
INV & $ \mbf M^{\text{INV}}( \mbf F ; \bs \theta) \cdot \mbf h(\mbf x)$ & $\begin{aligned}
\underset{\mbf V}{\text{solve}} \Big(& -\theta_1 \mbf K \mbf V 
+ \theta_2 \mbf A \mbf V \\ & -\theta_2 \mbf R \vdotop(\mbf V \otimes \mbf V \otimes \mbf V) 
- \mbf F = \mbf 0 \Big)
\end{aligned}$ & Known mechanisms & YES \\ \hline
SINDy & $ \mbf M^{\text{SINDy}}( \mbf F ; \bs \theta) \cdot \mbf h(\mbf x)$ &  $\underset{\mbf V}{\text{solve}} \Big(  \sum_{i} \theta_i \mbf L_i(\mbf V) + \mbf F = \mbf 0\Big)$& Possible mechanisms & YES \\ \hline
NBVP & $ \mbf M^{\text{NBVP}}( \mbf F ; \bs \theta) \cdot \mbf h(\mbf x)$ &  $\underset{\mbf V}{\text{solve}} \Big( \mbf N( \mbf V ; \bs \theta) + \mbf F = \mbf 0\Big)$& None & NO \\ \hline
NO & $ \mbf M^{\text{NO}}( \mbf F ; \bs \theta) \cdot \mbf h(\mbf x)$ & $\mbf M^{\text{NO}}(\mbf F ; \bs \theta)$ (neural network) & None & NO \\ \hline
\end{tabular}
\end{table}

\paragraph{} Table \ref{tab:summary} summarizes the results of the previous section. Given that each method can be written in the fixed basis model form, all the differences between them lie in the model class of the parameterized coefficient map. With the example of the static Allen-Cahn PDE, inverse problems parameterize known mechanisms in the PDE, whereas the SINDy library parameterized coefficients on a wide class of mechanisms derived from prior knowledge of the system of interest. These are considered to be mechanisms because they are informed by knowledge of the the ``activities'' of material points in a continuum system (e.g., relationships between the state field and the flux vector). Because the appropriate model class is enforced explicitly by the modeler, these models are capable of generalizing when the parameters are identifiable. In contrast, the neural BVP assumes that the differential equation is given by a neural network, and though the coefficients are still predicted through a nonlinear solve, we take this to be a phenomenological model. Unlike inverse problems and SINDy, the parameters of the neural network do not correspond clearly to mechanisms of the system of interest. Finally, neural operators forego the PDE model class, and thus the nonlinear solve, parameterizing the coefficient map as a deep neural network. Once again, the parameters do not have physical meaning and the model is not capable of generalizing. Though neural operators expedite the solution process by avoiding the nonlinear solve, they are phenomenological models that do not generalize.

\begin{figure}[hbt!]
\centering
\includegraphics[width=0.95\textwidth]{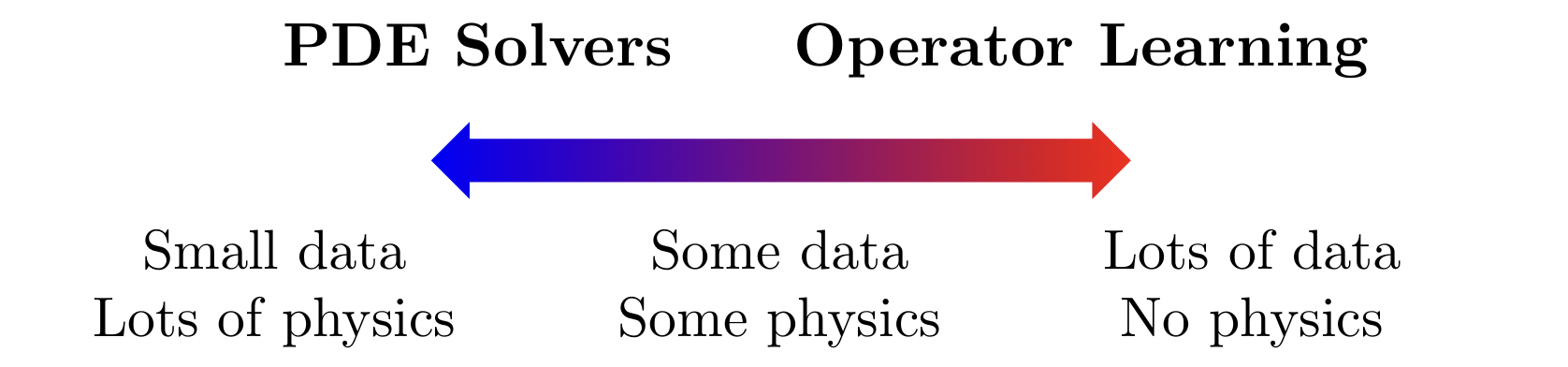}
\caption{In the scientific machine learning literature, modeling strategies are taxonomized by the availability of training data and prior physical knowledge. Image adapted from \cite{nicolas_boulle_overview_2024}.}
\label{spectrum}
\end{figure}

\begin{figure}[hbt!]
\centering
\includegraphics[width=0.7\textwidth]{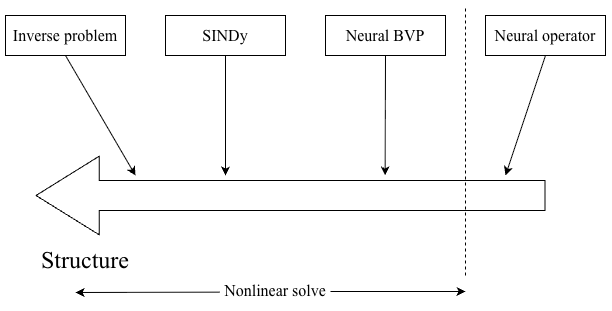}
\caption{We argue that the concept of structure is a more precise way to taxonomize different data-driven modeling strategies. Structure has to do with restrictions on the model class of the coefficient map. Generalization and the possibility of mechanism discovery boil down to nothing more than constraints on its mathematical form.}
\label{taxonomy}
\end{figure}

\begin{figure}[hbt!]
\centering
\includegraphics[width=0.6\textwidth]{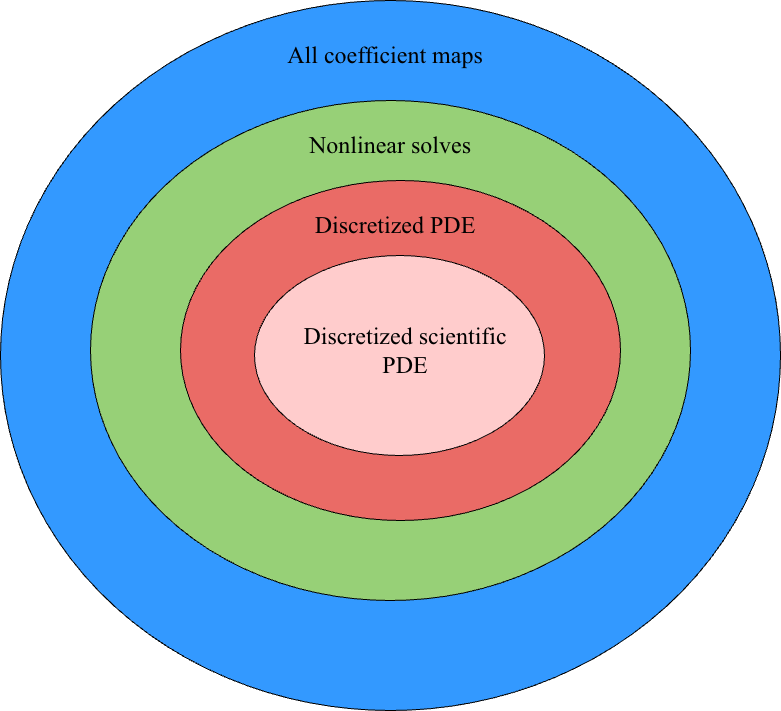}
\caption{Calling PDEs which are both parsimonious and simple ``scientific,'' we note that coefficient maps arising from their discretized solutions represent a small subset of all coefficient maps. Given that we have argued that scientific data comes from scientific PDEs, and that generalization is only possible when model classes are matched, there is a tension between the flexibility of the parametrized model and its ability to generalize.}
\label{scipde}
\end{figure}

\paragraph{} The taxonomy of data-driven models shown in Figure \ref{spectrum} will be familiar to the practitioner of data-driven modeling. In the literature, it is common to characterize neural operators and numerical solutions to PDEs as being two extremes of the data-physics spectrum. We agree with this basic framing, but find it to be insufficiently specific. What does it mean operationally to have ``knowledge of the physics''? In our framing, the difference between modeling strategies concentrates in the coefficient map, and prior knowledge restricts the form of the coefficient map. Knowledge of the physics appears in how discretized differential operators are parameterized in the nonlinear solve defining the relationship between the discretized forcing and the coefficients on the state. We find it more informative to taxonomize modeling strategies with the idea of ``structure,'' as this more clearly suggests building in constraints to the coefficient map. We take the notion of structure to encapsulate our discussions about model classes and mechanisms---a highly structured coefficient map is one that makes fine-grained assumptions about the model class of the data generating process, and thus mechanisms of the system of interest. We take the structure of the different modeling strategies to lie on a spectrum, and our proposed taxonomy is shown in Figure \ref{taxonomy}. Inverse problems are the most structured, in that the coefficient map comes from a discretized PDE solve built by a small number of parameters, all of which are expected to be non-zero. Though SINDy assumes the parsimonious/simple PDE model class, the governing equation can be built as a combination of extensive library terms. The method is thus less structured, as there are fewer constraints on the model class. Neural BVP foregoes the library and thus the parsimonious/simple model class, but retains the structure of a discretized PDE and nonlinear solve. Finally, neural operators are the least structured coefficient maps, as they rely neither on a governing equation not a nonlinear solve. While we grant that the data-physics taxonomy and our structured-based taxonomy agree on the ordering of data-driven modeling strategies, we think that the idea of structure more faithfully captures the differences between them. Finally, Figure \ref{scipde} emphasizes that coefficient maps arising from discretized PDEs which are both parsimonious and simple are a very small subset of all coefficient maps. This observation suggests a tension between the flexibility of the model (can it represent a wide array of behavior?) and the generalizability of the model (does it parameterize mechanisms?). We believe this observation to be both interesting and relevant for practitioners of scientific modeling.

%------------------------------------------------------------------------------

\section{Conclusion}

\paragraph{} With a static boundary value problem on a fixed geometry, we have argued that all models have a common form, whereby a coefficient map builds the state as a function of the input forcing with the help of basis functions. We then defined an optimization problem for the parameters of the coefficient map, and defined generalization to occur when the model continues to make accurate predictions outside of the convex training region. The entire model building process then centered on the parameterized coefficient map $\mbf M(\mbf F ; \bs \theta)$. Next, we asked the question: when does a learned model generalize? We argued that a model only generalizes when the parameters are identifiable and the model class matches that of the data generating process. We claimed that data on physical systems comes from the discretized PDEs with certain mathematical forms, e.g., comprising few terms and relatively benign nonlinearities. This meant that, at the very least, the coefficient map must be a nonlinear solve to generalize. We then connected to the philosophical literature on scientific models, arguing that when the mathematical form of the coefficient map (the model class) matched the data generating process (parsimonious and simple PDEs), the model could discover mechanisms underlying the data, and thus provide causal information about the system of interest. We took our conclusions about the model form and applied them to four data-driven modeling strategies in the literature, arguing that under the assumption that the data generating process is a PDE, only two of the four strategies we considered are capable of generalization. These ideas are certainly of interest in assessing the deployment of data-driven models in the real world. For example, we think it is misleading to claim that a neural operator ``learns'' the solution operator to a PDE, because, as a phenomenological model, it only summarizes and interpolates the data it was trained on. This means that one must guarantee that inputs to a neural operator used for a safety-critical engineering application never extrapolate the training data, as there are not a priori reasons to trust predictions in this regime. We note that there are some authors in the engineering literature pushing for data-driven ``foundation models'' which can generalize to inputs outside the training data \cite{choi_defining_2025}. Based on our analysis, we consider this to be impossible unless the governing differential equation is formulated and solved. Not surprisingly, given the emphasis on generalization in the cited work, the authors only use data to build a basis for the PDE solution, but still rely on a solve of the governing equation. As a final note, given the equivalence we draw between mechanisms and matching of model classes, and our argument that model class matching is a requirement for generalization, we were led to the conclusion that only mechanistic models generalize, in the sense that we have defined. This means that in order for a modeler to use the model to robustly reason about the system of interest---without careful consideration of whether model queries interpolate or extrapolate the training data---it is necessary for the model to describe entities (material points, conserved quantities) and activities (fluxes, sources) of the continuum system. We think clarity around this point is important in assessing the strengths and weaknesses of the various data-driven modeling strategies currently being investigated in the scientific machine learning literature.

%------------------------------------------------------------------------------

\appendix

%numbers equations in appendix
\counterwithin*{equation}{section}
\renewcommand\theequation{\thesection\arabic{equation}}

\section{Weak form of the Allen-Cahn equation}
\label{sec: allen-cahn}

\paragraph{} The governing PDE for the Allen-Cahn reaction-diffusion system is given by

\begin{equation*}
\begin{aligned}
    \kappa \nabla^2 \tilde u(\mbf x) + \sigma \tilde u(\mbf x)(1-\tilde u(\mbf x)^2) + f(\mbf x)=0, \quad \mbf x \in \Omega ,\\
    \tilde u(\mbf x) =0 , \quad \mbf x \in \partial \Omega,
\end{aligned}
\end{equation*}

\noindent where we take the boundaries to be zero for simplicity. Suppose that the output state is approximated with a finite set of basis function, e.g., $\tilde u(\mbf x) = \sum_{i=1}^{\mathcal B} \tilde U_i h_i(\mbf x)$, where the $h_i(\mbf x)$ are the basis functions. The goal of a numerical solution to the PDE is then to find the coefficients on the basis expansion of the output given the input $f(\mbf x)$. Using Galerkin projection, the coefficients can be found by plugging in the discretization to the governing equation, then enforcing that the PDE error is orthogonal to the space spanned by the basis functions. This is accomplished by enforcing orthogonality one basis function at a time:

\begin{equation*}
\begin{aligned}
    \kappa \int_{\Omega}  \sum_{i=1}^{\mathcal B} \tilde U_i \nabla^2 h_i(\mbf x) h_{\ell}(\mbf x)d\Omega + \sigma \int_{\Omega} \sum_{i=1}^{\mathcal B} \tilde U_i h_i(\mbf x) h_{\ell}(\mbf x)d\Omega - \sigma \int_{\Omega} \qty( \sum_{i=1}^{\mathcal B} \tilde U_i h_i(\mbf x))^3 h_{\ell}(\mbf x)d\Omega + \int_{\Omega} f(\mbf x) h_{\ell}(\mbf x)d\Omega = 0,
\end{aligned}
\end{equation*}

\noindent for $j=1,2,\dots,\mathcal B$. Pulling sums out of integrals, performing integration by parts on the first term, and expanding powers of the output field, we obtain

\begin{equation*}
\begin{aligned}
    -\kappa \sum_{i=1}^{\mathcal B} \tilde U_i \qty(\int_{\Omega} \nabla h_i \cdot \nabla h_{\ell} d\Omega) + \sigma \sum_{i=1}^{\mathcal B} \tilde U_i \qty(\int_{\Omega}  h_i h_{\ell} d\Omega) - \sigma \sum_{i=1}^{\mathcal B} \sum_{j=1}^{\mathcal B} \sum_{k=1}^{\mathcal B} \tilde U_i \tilde U_j \tilde U_k \qty(\int_{\Omega}  h_i h_j h_k h_{\ell}d\Omega) + \int_{\Omega} f h_{\ell} d\Omega = 0,
\end{aligned}
\end{equation*}

\noindent where the boundary term drops out of integration parts because, by assumption, the basis functions each satisfy $h_j(\mbf x)=0$ for $\mbf x \in \partial \Omega$. Defining vectors, matrices, and tensors from the integrated quantities in the above expression, we can write the governing system of equations in symbolic notation as 

\begin{equation*}
    -\kappa \mbf K \mbf{\tilde U} + \sigma \mbf A\mbf{\tilde U} - \sigma \mbf R \vdotop ( \mbf{\tilde U} \otimes \mbf{\tilde U} \otimes \mbf{\tilde U} ) + \mbf F = \mbf 0,
\end{equation*}

\noindent where symmetries of the four index tensor $\mbf R$ make it immaterial which indices are contracted on.

\section{Conservation of mass in a continuum system}
\label{sec: conservation}

\begin{figure}[hbt!]
\centering
\includegraphics[width=0.7\textwidth]{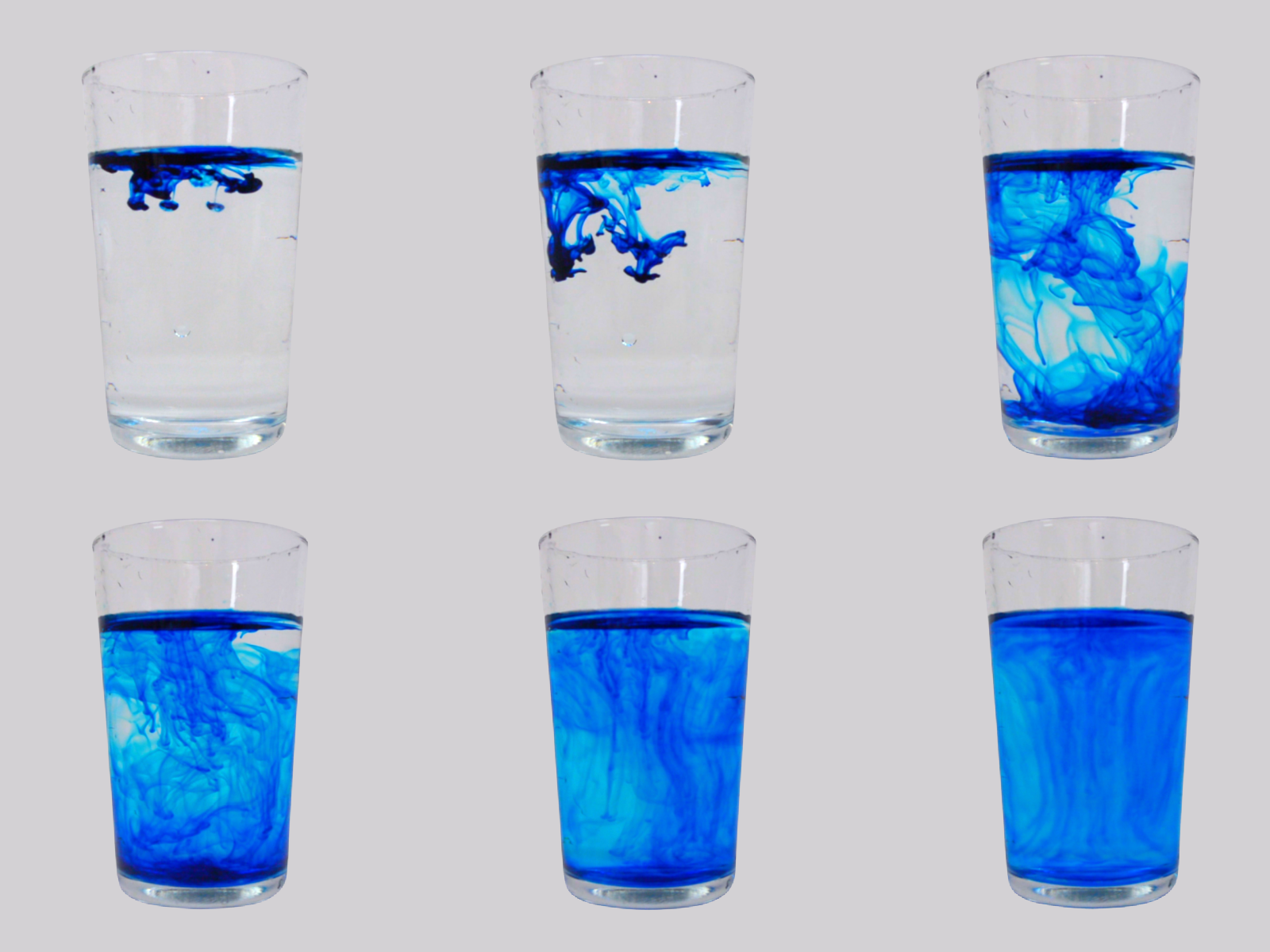}
\caption{The spreading of dye in a fluid medium is an example of a continuum system governed by the conservation of mass. Image adapted from \cite{vanstone_diffusion_2021}.}
\label{glass}
\end{figure}

\begin{figure}[hbt!]
\centering
\includegraphics[width=0.45\textwidth]{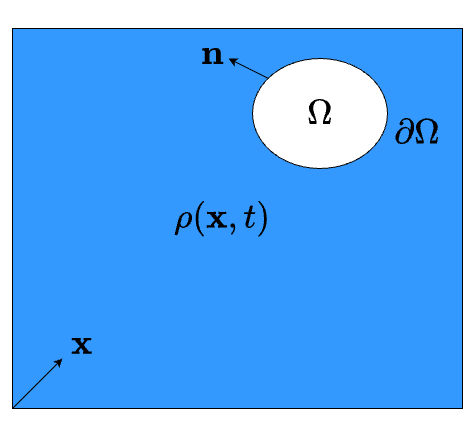}
\caption{A volume $\Omega$ with boundary $\partial \Omega$ immersed in a fluid medium. We are interested in the dynamics of the mixing of a dye in the fluid medium. This process can be modeled with the mass density $\rho(\mbf x,t)$ of the dye.}
\label{tank}
\end{figure}

\paragraph{} To illustrate the process of deriving a PDE model from a conservation law, consider the mass transport problem motivated by the process shown in Figure \ref{glass}. Blue dye is deposited at the top of a glass of water, and spreads to uniformly fill the cup over time. To model this process, we apply the conservation of mass to an arbitrary sub-volume of the system, which we call $\Omega$. We denote the mass density of dye $\rho(\mbf x,t)$, where $\mbf x$ is the spatial coordinate. As illustrated in Figure \ref{tank}, the boundary of the volume is denoted $\partial \Omega$, and the unit normal is $\mbf n$. Now,  the conservation of mass states that the rate of change of mass in the volume is due to explicit mass sources and the flow of mass through the boundary. Thus, we write

\begin{equation*}
   \underbrace{ \pd{}{t} \int_{\Omega} \rho(\mbf x, t ) d\Omega}_{\text{rate of change of mass}} = -\underbrace{ \int_{\partial \Omega} \mbf q(\rho) \cdot \mbf n dS}_{\text{mass flowing out}} + \underbrace{\int_{\Omega}f(\mbf x , t) d\Omega}_{\text{mass source term}},
\end{equation*}

\noindent where $\mbf q$ is called the ``mass flux.'' Passing the time derivative inside the volume integral and using the divergence theorem on the right-hand side, we can shrink the volume $\Omega$ down to a single point to obtain a governing PDE:

\begin{equation*}
    \underset{|\Omega|\rightarrow 0}{\text{lim }} \int_{\Omega} \qty(\pd{\rho}{t} + \nabla \cdot \mbf q(\rho) -f)d\Omega \implies \pd{\rho}{t} + \nabla \cdot \mbf q(\rho)- f = 0.
\end{equation*}

In order for this PDE to be solvable, we must specify how the mass flux depends on the density. Specifying the flux is a statement about what causes transport of the dye in the fluid medium. Connecting to the literature on mechanism, the entity at the scale of the mechanism is the density $\rho$ and the activity is the mass flux $\mbf q$. The mechanism is thus specified when the mass flux is written as a function of the density. For example, a common mechanism of mass transport is diffusion, which is written with Fick's law as

\begin{equation*}
    \underbrace{\mbf q^{\text{DIF}}(\rho)}_{\text{ mass flow}} = - \kappa \underbrace{ \nabla \rho(\mbf x, t)}_{\text{direction of increasing density}},
\end{equation*}

\noindent where $\kappa$ is an empirical property called the ``diffusivity'' parameterizing this mechanism of mass transport. The diffusivity, which we take to be constant in space and time, controls the mobility of mass in the fluid medium. Fick's law simply states the mass moves from high density to low density regions. As mentioned previously, we think of material properties as estimated from data by solving inverse problems, and thus as bona fide empirical constants. Another common mechanism of mass transport is advection, which is given by 

\begin{equation*}
    \underbrace{\mbf q^{\text{ADV}}(\rho)}_{\text{mass flow}} = \underbrace{\mbf v(\mbf x)  }_{\text{direction of fluid velocity field}}\rho(\mbf x),
\end{equation*}

\noindent where $\mbf v(\mbf x)$ is the velocity of the fluid medium at position $\mbf x$. The advective mechanism of mass transport is parameterized by this velocity field, which could be obtained by solving another PDE for the conservation of momentum in the fluid (itself parameterized by other material properties). Taking the mass transport mechanisms to be additive, we can write the governing equation for the density as

\begin{equation*}
    \pd{\rho}{t} + \nabla \cdot (\mbf q^{\text{DIF}} + \mbf q^{\text{ADV}})- f= \pd{\rho}{t} + \nabla \cdot(\mbf v \rho) - \kappa \nabla^2 \rho - f = 0.
\end{equation*}

This so-called ``advection-diffusion'' equation shows how PDEs from continuum mechanics decompose systems into infinitesimally small parts, comprising both entities and activities, and parameterize the activities, (mass flux, in this case), with empirical material properties. Again, we think of the PDE as implicitly defining a model class for the function mapping discretized forcings to the solution coefficients. As such, there is a strong connection between mechanistic models based on PDEs and the model class of nonlinear solves.

\section{Fully-connected neural network}
\label{sec: nn}

\paragraph{} In this Appendix, we show the mathematical form of a forward pass through a fully-connected deep neural network. Suppose that the input-output relation for the $i$-th hidden layer is

\begin{equation*}
    \mathbf{y}_i = \sigma\Big(  \mathbf{W}_i\mathbf{y}_{i-1} + \mathbf{b}_i  \Big),
\end{equation*}

\noindent where $\sigma(\cdot)$ is a nonlinear ``activation function'' applied element-wise, $\mbf W_i$ are the weights, and $\mbf b_i$ are the biases. As shown by the recursion above, the output $\mathbf{y}_i$ then becomes the next layer's input. This combination of an affine transformation of the previous layer and nonlinear activation is repeated for some user-specified number of layers. The parameters of the neural network are the collection of the weight matrices and bias vectors for each layer. Thus, we can write the neural network parameters as $\boldsymbol{\theta} = [ \mathbf{W}_1, \mathbf{b}_1, \mathbf{W}_2,\mathbf{b}_2,\dots]$. 

% As an example, the input is $\mbf y_0 = \mbf F \in \mathbb R^{\mathcal B}$ and the outputs are the solution coefficients $\mbf U \in \mathbb R^{\mathcal B}$. Thus, the neural network is simply a very flexible method to parameterize the coefficient map $\mbf M^{\text{NO}}: \mathbb R^{\mathcal B} \rightarrow \mathbb R^{\text{B}}$. 

\section{Weak form of Neural BVP}
\label{sec: nbvp}

\paragraph{} Computing a numerical solution to the differential operator represented by a neural network the weak form is very similar to a standard numerical PDE solution. With Neural BVPs, the governing equation is

\begin{equation*}
\begin{aligned}
    \mathcal N\qty( u ,  \pd{u}{x_1},\pd{u}{x_2},\pdd{u}{x_1},\pdd{u}{x_2},\dots; \bs \theta) + f(\mbf x) = 0 , \quad \mbf x \in \Omega, \\
    u(\mbf x) = 0 , \quad \mbf x \in \partial \Omega.
\end{aligned}
\end{equation*}

We discretize the output field with a finite set of basis functions that each satisfy the zero boundary conditions by construction. Plugging in the discretized solution, we again enforce that the error of the PDE is orthogonal to the space spanned by the basis functions. This reads

\begin{equation}
    \int_{\Omega} \mathcal N\qty( \sum_i U_i h_i, \sum_i U_i \pd{h_i}{x_1} , \sum_i U_i \pd{h_i}{x_2} , \sum_i U_i \pdd{h_i}{x_1} , \sum_i U_i \pdd{h_i}{x_2} ,\dots ; \bs \theta   ) h_j(\mbf x) d\Omega + \int_{\Omega} f(\mbf x) h_j(\mbf x)d\Omega = 0.
\end{equation}

For fixed model parameters $\bs \theta$, this is a nonlinear system of equations for the coefficients $\mbf U$. Unlike weak forms of standard PDEs, there are no simplifications or integrating by parts to be carried out. Thus, we write this nonlinear system as 

\begin{equation*}
    \mbf N(\mbf U ; \bs \theta ) + \mbf F = \mbf 0.
\end{equation*}

%------------------------------------------------------------------------------

% \bibliographystyle{unsrt}
% \bibliography{My_Library}

\begin{thebibliography}{10}

\bibitem{naser_what_2026}
Mohannad~Zeyad Naser.
\newblock What is an {Engineering} {Model}? {Philosophical} {Arguments} and {Counterarguments} on {Why} {Machine} {Learning} {Models} {Belong} in {Engineering}.
\newblock {\em European Journal for Philosophy of Science}, 16(2):46, 2026.

\bibitem{maxwell_ontological_1962}
Grover Maxwell.
\newblock The {Ontological} {Status} of {Theoretical} {Entities}.
\newblock In Herbert Feigl and Grover Maxwell, editors, {\em Scientific {Explanation}, {Space}, and {Time}: {Minnesota} {Studies} in the {Philosophy} of {Science}}, pages 181--192. University of Minnesota Press, 1962.

\bibitem{machamer_thinking_2000}
Peter Machamer, Lindley Darden, and Carl~F. Craver.
\newblock Thinking about {Mechanisms}.
\newblock {\em Philosophy of Science}, 67(1):1--25, 2000.

\bibitem{rohwer_how_2016}
Yasha Rohwer and Collin Rice.
\newblock How {Are} {Models} and {Explanations} {Related}?
\newblock {\em Erkenntnis}, 81(5):1127--1148, 2016.

\bibitem{kitcher_explanatory_1981}
Philip Kitcher.
\newblock {EXPLANATORY} {UNIFICATION}.
\newblock {\em Philosophy of Science}, 1981.

\bibitem{mckenna_laws_2026}
Travis McKenna.
\newblock Laws of {Nature} and {Their} {Supporting} {Casts}.
\newblock {\em British Journal for the Philosophy of Science}, 77(2):345--371, 2026.

\bibitem{weisberg_simulation_2013}
Michael Weisberg.
\newblock {\em Simulation and similarity : using models to understand the world}.
\newblock New York : Oxford University Press, 2013.

\bibitem{reddy_introduction_2013}
J.~N. Reddy.
\newblock {\em An {Introduction} to {Continuum} {Mechanics}}.
\newblock Cambridge University Press, Cambridge, 2 edition, 2013.

\bibitem{vallet_spectral_2008}
B.~Vallet and B.~Lévy.
\newblock Spectral {Geometry} {Processing} with {Manifold} {Harmonics}.
\newblock {\em Computer Graphics Forum}, 27(2):251--260, April 2008.

\bibitem{hughes_thomas_j_r_finite_2000}
Thomas J.~R. Hughes.
\newblock The {Finite} {Element} {Method}, 2000.

\bibitem{rowan_solving_2025}
Conor Rowan, Alireza Doostan, Kurt Maute, and John Evans.
\newblock Solving {Engineering} {Eigenvalue} {Problems} {With} {Neural} {Networks} {Using} the {Rayleigh} {Quotient}.
\newblock {\em International Journal for Numerical Methods in Engineering}, 126(24):e70209, 2025.
\newblock \_eprint: https://onlinelibrary.wiley.com/doi/pdf/10.1002/nme.70209.

\bibitem{li_fourier_2021}
Zongyi Li, Nikola Kovachki, Kamyar Azizzadenesheli, Burigede Liu, Kaushik Bhattacharya, Andrew Stuart, and Anima Anandkumar.
\newblock Fourier {Neural} {Operator} for {Parametric} {Partial} {Differential} {Equations}, May 2021.
\newblock arXiv:2010.08895 [cs].

\bibitem{ingebrand_basis--basis_2024}
Tyler Ingebrand, Adam~J. Thorpe, Somdatta Goswami, Krishna Kumar, and Ufuk Topcu.
\newblock Basis-to-{Basis} {Operator} {Learning} {Using} {Function} {Encoders}, November 2024.
\newblock arXiv:2410.00171 [cs].

\bibitem{tripura_wavelet_2023}
Tapas Tripura and Souvik Chakraborty.
\newblock Wavelet {Neural} {Operator} for solving parametric partial differential equations in computational mechanics problems.
\newblock {\em Computer Methods in Applied Mechanics and Engineering}, 404:115783, February 2023.

\bibitem{lu_comprehensive_2022}
Lu~Lu, Xuhui Meng, Shengze Cai, Zhiping Mao, Somdatta Goswami, Zhongqiang Zhang, and George~Em Karniadakis.
\newblock A comprehensive and fair comparison of two neural operators (with practical extensions) based on {FAIR} data.
\newblock {\em Computer Methods in Applied Mechanics and Engineering}, 393:114778, April 2022.

\bibitem{willcox_balanced_2002}
K.~Willcox and J.~Peraire.
\newblock Balanced {Model} {Reduction} via the {Proper} {Orthogonal} {Decomposition}.
\newblock {\em AIAA Journal}, 40(11):2323--2330, November 2002.

\bibitem{ghanem_stochastic_1991}
Roger~G. Ghanem and Pol~D. Spanos.
\newblock {\em Stochastic {Finite} {Elements}: {A} {Spectral} {Approach}}.
\newblock Springer, New York, NY, 1991.

\bibitem{raissi_physics-informed_2019}
M.~Raissi, P.~Perdikaris, and G.E. Karniadakis.
\newblock Physics-informed neural networks: {A} deep learning framework for solving forward and inverse problems involving nonlinear partial differential equations.
\newblock {\em Journal of Computational Physics}, 378:686--707, February 2019.

\bibitem{seidman_nomad_2022}
Jacob~H. Seidman, Georgios Kissas, Paris Perdikaris, and George~J. Pappas.
\newblock {NOMAD}: {Nonlinear} {Manifold} {Decoders} for {Operator} {Learning}, June 2022.
\newblock arXiv:2206.03551 [cs].

\bibitem{winsberg_science_2010}
Eric Winsberg.
\newblock {\em Science in the {Age} of {Computer} {Simulation}}.
\newblock University of Chicago Press, Chicago, IL, October 2010.

\bibitem{herfel_theories_2023}
William~E. Herfel, Władysław Krajewski, Ilkka Niiniluoto, and Ryszard Wójcicki.
\newblock Theories and {Models} in {Scientific} {Processes}.
\newblock In {\em Theories and {Models} in {Scientific} {Processes}}. Brill, December 2023.

\bibitem{morgan_models_1999}
Mary~S. Morgan and Margaret Morrison, editors.
\newblock {\em Models as {Mediators}: {Perspectives} on {Natural} and {Social} {Science}}.
\newblock Ideas in {Context}. Cambridge University Press, Cambridge, 1999.

\bibitem{morrison_reconstructing_2015}
Margaret Morrison.
\newblock {\em Reconstructing {Reality}: {Models}, {Mathematics}, and {Simulations}}.
\newblock Oxford {Studies} in {Philosophy} of {Science}. Oxford University Press, Oxford, New York, January 2015.

\bibitem{mcmullin_what_1968}
E.~McMullin.
\newblock What do {Physical} {Models} {Tell} us?
\newblock 52:385--396, 1968.
\newblock Book Title: Studies in Logic and the Foundations of Mathematics ISBN: 9780444534163.

\bibitem{termine_machine_2026}
Alberto Termine, Emanuele Ratti, and Alessandro Facchini.
\newblock Machine {Learning} and {Theory} {Ladenness} -- {A} {Phenomenological} {Account}, January 2026.
\newblock arXiv:2409.11277 [cs].

\bibitem{giere_how_2004}
Ronald~N. Giere.
\newblock How {Models} {Are} {Used} to {Represent} {Reality}.
\newblock {\em Philosophy of Science}, 71(5):742--752, December 2004.

\bibitem{frigg_models_2021}
Roman Frigg and Stephan Hartmann.
\newblock Models in {Science} (2nd edition).
\newblock In Edward~N. Zalta, editor, {\em The {Stanford} {Encyclopedia} of {Philosophy}}. Stanford, 2021.

\bibitem{van_fraassen_imaging_2008}
Bas~C. van Fraassen.
\newblock Imaging, {Picturing}, and {Scaling}.
\newblock In Bas~C. van Fraassen, editor, {\em Scientific {Representation}: {Paradoxes} of {Perspective}}, page~0. Oxford University Press, August 2008.

\bibitem{nguyen_scientific_2022}
James Nguyen and Roman Frigg.
\newblock {\em Scientific {Representation}}.
\newblock Elements in the {Philosophy} of {Science}. Cambridge University Press, Cambridge, 2022.

\bibitem{craver_when_2006}
Carl~F. Craver.
\newblock When {Mechanistic} {Models} {Explain}.
\newblock {\em Synthese}, 153(3):355--376, 2006.

\bibitem{siegel_phenomenological_2024}
Gabriel Siegel and Carl~F. Craver.
\newblock Phenomenological {Laws} and {Mechanistic} {Explanations}.
\newblock {\em Philosophy of Science}, 91(1):132--150, January 2024.

\bibitem{povich_mechanistic_2017}
Mark Povich and Carl~F. Craver.
\newblock Mechanistic {Levels}, {Reduction}, and {Emergence}.
\newblock In Stuart Glennan and Phyllis Illari, editors, {\em The {Routledge} {Handbook} of {Mechanisms} and {Mechanical} {Philosophy}}, pages 185--97. Routledge, 2017.

\bibitem{ioannidis_mechanisms_2018}
Stavros Ioannidis and Stathis Psillos.
\newblock Mechanisms in practice: {A} methodological approach.
\newblock {\em Journal of Evaluation in Clinical Practice}, 24(5):1177--1183, October 2018.

\bibitem{suarez_inferential_2004}
Mauricio Suárez.
\newblock An {Inferential} {Conception} of {Scientific} {Representation}.
\newblock {\em Philosophy of Science}, 71(5):767--779, December 2004.

\bibitem{bueno_inferential_2011}
Otávio Bueno and Mark Colyvan.
\newblock An {Inferential} {Conception} of the {Application} of {Mathematics}.
\newblock {\em Noûs}, 45(2):345--374, 2011.

\bibitem{khalifa_scientific_2022}
Kareem Khalifa, Jared Millson, and Mark Risjord.
\newblock Scientific {Representation}: {An} {Inferentialist}-{Expressivist} {Manifesto}.
\newblock {\em Philosophical Topics}, 50(1):263--291, 2022.

\bibitem{woodward_what_2002}
James Woodward.
\newblock What is a {Mechanism}? {A} {Counterfactual} {Account}.
\newblock {\em Philosophy of Science}, 69(S3):366--377, 2002.

\bibitem{bretin_penalized_2024}
Elie Bretin, Chih-Kang Huang, and Simon Masnou.
\newblock A penalized {Allen}-{Cahn} equation for the mean curvature flow of thin structures, May 2024.
\newblock arXiv:2310.10272 [math].

\bibitem{irving_statistical_1950}
J.~H. Irving and John~G. Kirkwood.
\newblock The {Statistical} {Mechanical} {Theory} of {Transport} {Processes}. {IV}. {The} {Equations} of {Hydrodynamics}.
\newblock {\em Journal of Chemical Physics}, 18:817--829, June 1950.
\newblock ADS Bibcode: 1950JChPh..18..817I.

\bibitem{brunton_discovering_2016}
Steven~L. Brunton, Joshua~L. Proctor, and J.~Nathan Kutz.
\newblock Discovering governing equations from data by sparse identification of nonlinear dynamical systems.
\newblock {\em Proceedings of the National Academy of Sciences}, 113(15):3932--3937, April 2016.

\bibitem{messenger_weak_2021}
Daniel~A. Messenger and David~M. Bortz.
\newblock Weak {SINDy} {For} {Partial} {Differential} {Equations}.
\newblock {\em Journal of Computational Physics}, 443:110525, October 2021.
\newblock arXiv:2007.02848 [math].

\bibitem{udrescu_ai_2020}
Silviu-Marian Udrescu and Max Tegmark.
\newblock {AI} {Feynman}: a {Physics}-{Inspired} {Method} for {Symbolic} {Regression}, April 2020.
\newblock arXiv:1905.11481 [physics].

\bibitem{cranmer_interpretable_2023}
Miles Cranmer.
\newblock Interpretable {Machine} {Learning} for {Science} with {PySR} and {SymbolicRegression}.jl, May 2023.
\newblock arXiv:2305.01582 [astro-ph].

\bibitem{wigner_unreasonable_1960}
Eugene Wigner.
\newblock The {Unreasonable} {Effectiveness} of {Mathematics} in the {Natural} {Sciences}.
\newblock {\em Communications on Pure and Applied Mathematics}, 1960.

\bibitem{fraassen_arguments_1998}
Bas~Van Fraassen.
\newblock Arguments {Concerning} {Scientific} {Realism}.
\newblock In Martin Curd and Jan~A. Cover, editors, {\em Philosophy of {Science}: {The} {Central} {Issues}}. Norton, 1998.

\bibitem{shea_sindy-bvp_2021}
Daniel~E. Shea, Steven~L. Brunton, and J.~Nathan Kutz.
\newblock {SINDy}-{BVP}: {Sparse} identification of nonlinear dynamics for boundary value problems.
\newblock {\em Physical Review Research}, 3(2):023255, June 2021.

\bibitem{zhang_understanding_2017}
Chiyuan Zhang, Samy Bengio, Moritz Hardt, Benjamin Recht, and Oriol Vinyals.
\newblock Understanding deep learning requires rethinking generalization, February 2017.
\newblock arXiv:1611.03530 [cs].

\bibitem{huang_understanding_2020}
W.~Ronny Huang, Zeyad Emam, Micah Goldblum, Liam Fowl, J.~K. Terry, Furong Huang, and Tom Goldstein.
\newblock Understanding {Generalization} through {Visualizations}, November 2020.
\newblock arXiv:1906.03291 [cs].

\bibitem{belkin_reconciling_2019}
Mikhail Belkin, Daniel Hsu, Siyuan Ma, and Soumik Mandal.
\newblock Reconciling modern machine-learning practice and the classical bias–variance trade-off.
\newblock {\em Proceedings of the National Academy of Sciences}, 116(32):15849--15854, August 2019.

\bibitem{luxburg_statistical_2011}
Ulrike~von Luxburg and Bernhard Schölkopf.
\newblock Statistical {Learning} {Theory}: {Models}, {Concepts}, and {Results}.
\newblock In Dov~M. Gabbay, Stephan Hartmann, and John Woods, editors, {\em Inductive {Logic}}, volume~10 of {\em Handbook of the {History} of {Logic}}, pages 651--706. North-Holland, January 2011.

\bibitem{wieland_structural_2021}
Franz-Georg Wieland, Adrian~L. Hauber, Marcus Rosenblatt, Christian Tönsing, and Jens Timmer.
\newblock On structural and practical identifiability.
\newblock {\em Current Opinion in Systems Biology}, 25:60--69, March 2021.

\bibitem{maclaren_what_2020}
Oliver~J. Maclaren and Ruanui Nicholson.
\newblock What can be estimated? {Identifiability}, estimability, causal inference and ill-posed inverse problems, July 2020.
\newblock arXiv:1904.02826 [math].

\bibitem{bourdin_numerical_2000}
B.~Bourdin, G.~A. Francfort, and J-J. Marigo.
\newblock Numerical experiments in revisited brittle fracture.
\newblock {\em Journal of the Mechanics and Physics of Solids}, 48(4):797--826, April 2000.

\bibitem{craver_ontic_2014}
Carl~F. Craver.
\newblock The {Ontic} {Account} of {Scientific} {Explanation}.
\newblock In Marie~I. Kaiser, Oliver~R. Scholz, Daniel Plenge, and Andreas Hüttemann, editors, {\em Explanation in the special science: {The} case of biology and history}, pages 27--52. Springer, 2014.

\bibitem{glennan_rethinking_2002}
Stuart Glennan.
\newblock Rethinking {Mechanistic} {Explanation}.
\newblock {\em Philosophy of Science}, 69(S3):342--353, 2002.

\bibitem{bechtel_explanation_2005}
William Bechtel and Adele Abrahamsen.
\newblock Explanation: a mechanist alternative.
\newblock {\em Studies in History and Philosophy of Biological and Biomedical Sciences}, 36(2):421--441, June 2005.

\bibitem{dowe_wesley_1992}
Phil Dowe.
\newblock Wesley {Salmon}?s {Process} {Theory} of {Causality} and the {Conserved} {Quantity} {Theory}.
\newblock {\em Philosophy of Science}, 59(2):195--216, 1992.

\bibitem{sargsyan_statistical_2015}
Khachik Sargsyan, Roger Ghanem, and HN~Najm.
\newblock On the {Statistical} {Calibration} of {Physical} {Models}.
\newblock {\em International Journal of Chemical Kinetics}, 2015.

\bibitem{rowan_extending_2025}
Conor Rowan.
\newblock Extending the explicit constraint force method to inverse problems, December 2025.
\newblock arXiv:2512.14877 [math].

\bibitem{rudy_data-driven_2016}
Samuel~H. Rudy, Steven~L. Brunton, Joshua~L. Proctor, and J.~Nathan Kutz.
\newblock Data-driven discovery of partial differential equations, September 2016.
\newblock arXiv:1609.06401 [nlin].

\bibitem{donoho_compressed_2006}
DL~Donoho.
\newblock Compressed {Sensing}.
\newblock 52:1289--1306, April 2006.

\bibitem{chen_neural_2019}
Ricky T.~Q. Chen, Yulia Rubanova, Jesse Bettencourt, and David Duvenaud.
\newblock Neural {Ordinary} {Differential} {Equations}, December 2019.
\newblock arXiv:1806.07366 [cs].

\bibitem{wang_learning_2024}
Cong Wang, Aoming Liang, Fei Han, Xinyu Zeng, Zhibin Li, Dixia Fan, and Jens Kober.
\newblock Learning {Adaptive} {Hydrodynamic} {Models} {Using} {Neural} {ODEs} in {Complex} {Conditions}, October 2024.
\newblock arXiv:2410.00490 [cs].

\bibitem{sorourifar_physics-enhanced_2023}
Farshud Sorourifar, You Peng, Ivan Castillo, Linh Bui, Juan Venegas, and Joel~A. Paulson.
\newblock Physics-{Enhanced} {Neural} {Ordinary} {Differential} {Equations}: {Application} to {Industrial} {Chemical} {Reaction} {Systems}.
\newblock {\em Industrial \& Engineering Chemistry Research}, 62(38):15563--15577, September 2023.

\bibitem{lu_deeponet_2021}
Lu~Lu, Pengzhan Jin, and George~Em Karniadakis.
\newblock {DeepONet}: {Learning} nonlinear operators for identifying differential equations based on the universal approximation theorem of operators.
\newblock {\em Nature Machine Intelligence}, 3(3):218--229, March 2021.
\newblock arXiv:1910.03193 [cs].

\bibitem{cao_lno_2023}
Qianying Cao, Somdatta Goswami, and George~Em Karniadakis.
\newblock {LNO}: {Laplace} {Neural} {Operator} for {Solving} {Differential} {Equations}, May 2023.
\newblock arXiv:2303.10528 [cs].

\bibitem{hao_laplacian_2025}
Wenrui Hao and Jindong Wang.
\newblock Laplacian {Eigenfunction}-{Based} {Neural} {Operator} for {Learning} {Nonlinear} {Partial} {Differential} {Equations}, February 2025.
\newblock arXiv:2502.05571 [math-ph].

\bibitem{lanthaler_operator_2023}
Samuel Lanthaler.
\newblock Operator learning with {PCA}-{Net}: upper and lower complexity bounds, October 2023.
\newblock arXiv:2303.16317 [cs].

\bibitem{zhang_finite_2025}
Zecheng Zhang, Hao Liu, Guosheng Fu, Hayden Schaeffer, and Guang Lin.
\newblock Finite {Element} {Representation} {Network} ({FERN}) for {Operator} {Learning} with a {Localized} {Trainable} {Basis}, October 2025.
\newblock arXiv:2510.26962 [math].

\bibitem{nicolas_boulle_overview_2024}
An overview of operator learning, October 2024.

\bibitem{choi_defining_2025}
Youngsoo Choi, Siu~Wun Cheung, Youngkyu Kim, Ping-Hsuan Tsai, Alejandro~N. Diaz, Ivan Zanardi, Seung~Whan Chung, Dylan~Matthew Copeland, Coleman Kendrick, William Anderson, Traian Iliescu, and Matthias Heinkenschloss.
\newblock Defining {Foundation} {Models} for {Computational} {Science}: {A} {Call} for {Clarity} and {Rigor}, May 2025.
\newblock arXiv:2505.22904 [cs.LG].

\bibitem{vanstone_diffusion_2021}
Emma Vanstone.
\newblock Diffusion {Demonstration}, March 2021.

\end{thebibliography}

\end{document}